%% file: main.tex
\documentclass{article} 
\usepackage{iclr2027_conference,times}

\input{math_commands.tex}

\usepackage{hyperref}
\hypersetup{hidelinks}
\usepackage{url}

\input{macro}

\title{Learning Words by Simulating Worlds}

\input{authors}

\iclrfinalcopy 

\renewcommand{\headrulewidth}{0pt}

\begin{document}

\maketitle

\input{sec/00-abstract}
\input{sec/01-introduction}
\input{sec/03-method}
\input{sec/04-setup}
\input{sec/05-findings}
\input{sec/07-ablation}
\input{sec/02-background}
\input{sec/06-conclusion}

\section*{AI use statement}
We used generative AI tools for dataset construction and annotation, including the generation and filtering of synthetic action--state transition data, and for assistance with software implementation and code development.
All AI-assisted code was reviewed and tested by the authors, and the resulting datasets were subject to author-designed filtering and quality-control procedures.

We did not use generative AI to develop the core conceptual framework, formulate or refine the research hypotheses, design the research methodology
or experiments, formulate mathematical claims or proofs, or interpret the experimental results; these decisions and analyses were made by the authors.

Additionally, we used generative AI for final-stage language and grammar editing to improve the readability of the manuscript. All AI-assisted edits were reviewed by the authors. We take responsibility for the final content of this work, including its data, code, analyses, claims, and text.

\bibliography{references}
\bibliographystyle{iclr2027_conference}

\newpage
\appendix
\input{appendix/appendix}

\end{document}

%% file: math_commands.tex
\usepackage{amsmath,amsfonts,bm}

\def\eqref#1{equation~\ref{#1}}

\def\1{\bm{1}}

\DeclareMathAlphabet{\mathsfit}{\encodingdefault}{\sfdefault}{m}{sl}
\SetMathAlphabet{\mathsfit}{bold}{\encodingdefault}{\sfdefault}{bx}{n}



%% file: macro.tex
\usepackage{times}
\usepackage{latexsym}

\usepackage[T1]{fontenc}

\usepackage[utf8]{inputenc}

\usepackage{microtype}

\usepackage{inconsolata}

\usepackage{graphicx}

\usepackage{hyperref}       
\usepackage{url}            
\usepackage{booktabs}       
\usepackage{amsfonts}       
\usepackage{nicefrac}       
\usepackage{microtype}      
\usepackage{xcolor}         

\usepackage{rotating}

\usepackage[most]{tcolorbox}
\tcbuselibrary{listings,breakable}

\usepackage{algorithm}
\usepackage{algpseudocode}
\usepackage{amsmath}
\usepackage{algorithmicx}
\usepackage{booktabs}
\usepackage{multirow}
\usepackage{multicol}
 \usepackage{mathtools}
\usepackage{amsfonts}
\usepackage{cleveref}
\usepackage{pdfcol}
\usepackage{pdfrender}
\usepackage{wrapfig,lipsum}
\usepackage{tabularx}
\usepackage{array}
\usepackage{makecell}

\usepackage{subcaption}

\usepackage[shortlabels]{enumitem}
\setlist[itemize]{leftmargin=*}
\setlist[enumerate]{leftmargin=*}
\usepackage[normalem]{ulem}

\definecolor{commentgrey}{gray}{0.7}

\usepackage{todonotes}

\newcommand{\sota}[1]{%
  \text{\bfseries
    \pdfrender{StrokeColor=black,TextRenderingMode=2,LineWidth=0.3pt}{#1}}%
}

\usepackage{subcaption}

\newlength{\simactcellwL}          
\newlength{\simactcellwR}          
\newlength{\simactpanelh}
\newcommand{\simactcelltune}[6]{%
    \parbox[c][\simactpanelh][c]{#6}{%
        \centering
        \raisebox{#4}[\height][\depth]{%
            \includegraphics[
                width=#2#6,
                height=#3\simactpanelh,
                keepaspectratio,
                trim=#5,
                clip
            ]{#1}%
        }%
    }%
}

\newcommand{\simactheader}{%
    \parbox[c]{\simactcellwL}{\centering\footnotesize\bfseries Noun Learning Trajectory}\hfill
    \parbox[c]{\simactcellwL}{\centering\footnotesize\bfseries Verb Learning Trajectory}\hfill
    \parbox[c]{\simactcellwR}{\centering\footnotesize\bfseries Pre-condition Image}\hfill
    \parbox[c]{\simactcellwR}{\centering\footnotesize\bfseries Post-condition Image}%
    \par\vspace*{-6pt}%
}

\newcommand{\simactrowcustom}[6]{%
    \begin{subfigure}[t]{\textwidth}
        \centering
        #1\hfill
        #2\hfill
        #3\hfill
        #4
        \vspace*{-10pt}
        \caption{#5}
        \label{#6}
        \vspace*{-5pt}
    \end{subfigure}
    
}

\newtcblisting{promptbox}[1]{%
    breakable,
    width=\textwidth,
    colback=gray!3!white,
    colframe=gray!60!black,
    title={#1},
    fonttitle=\bfseries,
    listing only,
    listing engine=listings,
    listing options={
        basicstyle=\small\ttfamily,
        breaklines=true,
        columns=fullflexible,
        keepspaces=true,
        showstringspaces=false,
        breakatwhitespace=false,
        literate=
          {—}{{---}}1
          {→}{{$\rightarrow$}}1
    },
    before skip=6pt,
    after skip=6pt
}

%% file: authors.tex
\newcommand{\umich}{$^{1}$}
\newcommand{\upenn}{$^{2}$}
\newcommand{\amz}{$^{3}$}
\newcommand{\uwat}{$^{4}$}
\newcommand{\vect}{$^{5}$}
\newcommand{\mitech}{$^{6}$}

\newcommand{\andaffto}{$^,$}

\author{%
    Ziqiao Ma\umich\thanks{Corresponding author: \texttt{marstin@umich.edu}}\quad
    Jing Ding\umich\quad
    Sihan Xu\umich\quad
    Stella X. Yu\umich\quad
    Jiayuan Mao\upenn\andaffto\amz\quad \\
    {\bf Freda Shi\uwat\andaffto\vect\quad
    Joshua B Tenenbaum\mitech\quad
    Joyce Chai\umich}  \\
    \umich University of Michigan \quad
    \upenn University of Pennsylvania \quad
    \amz Amazon FAR \\
    \uwat University of Waterloo \quad
    \vect Vector Institute \quad
    \mitech Massachusetts Institute of Technology
}

%% file: sec/00-abstract.tex
\begin{abstract}

Result verbs such as \textit{open}, \textit{slice}, and \textit{fill} describe not only what is present in a scene, but how an action transforms one world state into another.
We study whether this causal structure provides a privileged learning signal for language models.
We train an autoregressive model on interleaved language and visual transition sequences, where each example contains a triple of a pre-condition image, an action description, and a post-condition image.
The model is trained with complementary objectives: inferring the action from an observed pre/post transition, predicting the post-condition from the pre-condition and action, and reconstructing the pre-condition from the post-condition and action.
Across controlled ablations, visual transition simulation improves grounded word acquisition when it is paired with sufficient language supervision.
The gains are concentrated on verbs rather than nouns, and within verbs, on result verbs rather than manner verbs, consistent with the fact that result verbs are defined by observable state changes while manner verbs depend more on motion trajectories.
These findings suggest that grounding verb semantics requires more than aligning words with percepts: it requires predictive models of how the world changes.

\end{abstract}

%% file: sec/01-introduction.tex
\section{Introduction}

Children cannot understand and learn the verb \textit{open} from a single image, as the word denotes a transition over time: something is closed, an action occurs, and the same thing becomes open.
Many concrete verbs have this structure: they do not merely name what is visible in a scene; instead, they describe how agents, objects, and states are related through changes.
To learn such words, a learner must perceive not only objects, but also the causal transformations in which those objects participate.

This observation raises a question for grounded language acquisition.
Most benchmarks and methods for visual grounding treat language as a mapping from words to percepts: nouns to objects~\citep{xu2007word}, adjectives to attributes~\citep{buettner2024investigating}, and referring expressions to regions~\citep{kazemzadeh2014referitgame,yu2016modeling}.
This formulation has been highly effective for object-centered grounding, but it is incomplete for actions~\citep{ma2023world}.
Verbs such as \textit{open}, \textit{slice}, or \textit{fill} are not grounded in a single region or frame.
Their meanings are distributed across an initial state (pre-condition) and a resulting state (post-condition).
The perceptual evidence is not only what is present, but what has changed.

This distinction also exposes a limitation in contemporary Vision-Language Models \citep[VLMs;][\textit{inter alia}]{Gemini2023,gpt4o,anthropic2024claude}.
Although VLMs can describe objects, actions, and events, they are still largely grounded through recognition, image-caption alignment, and instruction following: perception is used to answer prompts, not to model how the world changes~\citep{gao2025vision}.
This is insufficient as grounding an action verb requires representing the causal transition it denotes: the state before the action, the state after it, and the relation by which one becomes the other~\citep{gao2018causality}.
Such transition-centered representations are the basis of \textit{world models}, which support prediction, monitoring, and planning by simulating how actions transform states~\citep{wooldridge1995intelligent,ha2018world,lecun2022path}.
From this perspective, current VLM failures on spatiotemporal perception, intuitive physics, future-state prediction, and agentic causality are not merely benchmark errors, but failures of grounded world modeling~\citep{goyal2020rel3dminimallycontrastivebenchmark,shangguan2024tomatoassessingvisualtemporal,fu2024blink,zhang2025do,bakhtin2019phyrenewbenchmarkphysical,yi2020clevrer,bear2022physion}.
They motivate recent calls to move beyond reactive perception and brute-force long context toward models that construct, update, and predict with implicit world models from continual sensory experience~\citep{yang2025cambrians}.

We see this as a grounded word acquisition problem.
Our hypothesis is specific: for concrete predicates whose meanings are partly defined by observable state change, action-conditioned visual prediction should provide a privileged learning signal.
If a verb is individuated by the transition it licenses, then predicting the resulting state from the initial state and the action should expose the learner to the variables that matter for that verb: the participants, their initial configuration, their resulting configuration, and the causal relation between them.
This hypothesis does not predict that visual prediction should uniformly improve all words.
It predicts a structured dissociation: {\em transition prediction should benefit verbs more than nouns, with the strongest gains for result verbs, whose meanings are defined by the states they bring about, rather than manner verbs, which depend more on how an action unfolds over time}~\citep{levin2008constraint}.

We test this hypothesis in a modern autoregressive modeling setup, training on interleaved language and visual transition sequences to instantiate a joint autoregressive world-language model.
Each example contains a pre-condition image, an action description, and a post-condition image.
The model is trained with three complementary objectives: transition captioning (inferring the action from the pre-condition and post-condition), forward simulation (predicting the post-condition from the pre-condition and action), and backward simulation (inferring the pre-condition from the post-condition and action).
These objectives place language prediction and visual state prediction in the same representational system, allowing us to ask whether simulation-like training changes the acquisition dynamics of concrete words.

Our results support this hypothesis.
Action-conditioned simulation improves lexical acquisition, but only when paired with sufficient language-side supervision.
Its benefits are not uniform: they are concentrated on verbs rather than nouns, and within verbs, on result verbs rather than manner verbs.
This dissociation reflects the structure of the world-modeling signal: before-and-after images make changes of state directly observable, providing evidence for how actions transform the world, but only weakly encode how an action unfolds over time.
Accordingly, simulation helps most when a word's meaning is defined by the outcome it brings about.
These findings suggest that grounding is not merely mapping words to percepts; for verbs, it requires joint world-language modeling, where linguistic forms are grounded to the predictive representations of how actions transform the world.

%% file: sec/03-method.tex
\section{Learning Language by Simulation}

\input{floats/model-and-data}

We minimally formulate neural word learning from visual simulation as a single autoregressive sequence model (Figure~\ref{fig:method}). 
Each training example is represented as an interleaved sequence of visual and textual units: a pre-condition image, an action description, and a post-condition image. 
A standard causal Transformer processes the sequence from left to right and is trained to predict the next element at each position. For text positions, the model uses the usual next-token prediction objective. 
For visual positions, instead of predicting pixels directly, the model predicts the next visual embedding, allowing visual states to be modeled in the same autoregressive framework as language.

\subsection{Model Architecture and Implementations}

We adopt the assumption that word learning can rely on a visual prior but not on a language prior, such as a pretrained language model. 
Accordingly, we initialize the visual backbone from NEPA~\citep{xu2025next}, an autoregressive visual model pretrained with causal patch-embedding prediction, while initializing the language-specific components from scratch. 
We use whole-word tokenization so that prediction units correspond to the lexical units under study.
A language head tied to the text embedding matrix \citep{radford2019language} predicts the next word, while a linear image head predicts the next patch embedding. 
Architectural dimensions, positional encodings, and component initialization are detailed in Appendix~\ref{app:architecture}.




\subsection{Sequence Formats}
\label{sec:sequence-format}

\input{tables/sequence-format}

Each training example contains a pre-condition image, an action text, and a post-condition image. 
At load time, we sample one of four sequence formats with fixed probabilities (Table~\ref{tab:sequence_formats}). 
All formats share the same model parameters and the same causal next-element prediction objective. 
The supervision regime determines which positions contribute to the loss (Section~\ref{sec:supervision_regime}). 
Serialization and sampling details in Appendix~\ref{app:sequence}.

\subsection{Training Objectives}
\label{sec:training-objective}

The total loss combines supervised next-token prediction for text positions and supervised next-embedding prediction for image positions:
\vspace*{-4pt}
\begin{equation}
\mathcal{L}
=
\mathcal{L}_{\mathrm{LM}}
+
\mathcal{L}_{\mathrm{SIM}} .
\vspace*{-4pt}
\end{equation}
Both terms are cross-entropy losses: $\mathcal{L}_{\mathrm{LM}}$ is cross-entropy over tokens, while $\mathcal{L}_{\mathrm{SIM}}$ is a visual simulation loss over patch embeddings.

\textbf{Language modeling loss.}
For each supervised text position, the Transformer hidden state $h_t$ is passed through the language head to predict the next whole-word token:
\vspace*{-4pt}
\begin{equation}
p_{\theta}(x_{t+1} \mid x_{\leq t})
=
\mathrm{softmax}(\mathrm{lm\_head}(h_t)).
\vspace*{-4pt}
\end{equation}
The language modeling loss is standard causal cross-entropy over the supervised text positions:
\vspace*{-4pt}
\begin{equation}
\mathcal{L}_{\mathrm{LM}}
=
-\frac{1}{|\mathcal{S}_{\mathrm{text}}|}
\sum_{t \in \mathcal{S}_{\mathrm{text}}}
\log p_{\theta}(x_{t+1} \mid x_{\leq t}).
\vspace*{-4pt}
\end{equation}
The set $\mathcal{S}_{\mathrm{text}}$ is determined by the sampled sequence format and supervision regime.

\textbf{Image prediction loss.}
For each supervised image position, the model predicts the next patch embedding from the current hidden state:
\vspace*{-4pt}
\begin{equation}
\hat{z}_{t+1} = \mathrm{img\_head}(h_t),
\vspace*{-4pt}
\end{equation}
The target embedding $z_{t+1}$ is produced by the same patch embedding module applied to the corresponding image patch.
Our main image objective is an InfoNCE loss over supervised image positions. 
This keeps the language and image objectives on comparable scales without introducing additional loss weight.
Given all supervised image predictions and targets in a batch, we $\ell_2$-normalize both sets of embeddings and compute the similarity matrix
\vspace*{-4pt}
\begin{equation}
s_{ij}
=
\hat{z}_i^{\top} z_j / \tau,
\vspace*{-4pt}
\end{equation}
where $\tau$ is the temperature. 
The image loss is a cross-entropy loss that identifies the correct target embedding for each predicted embedding:
\vspace*{-4pt}
\begin{equation}
\mathcal{L}_{\mathrm{SIM}}
=
-\frac{1}{|\mathcal{S}_{\mathrm{SIM}}|}
\sum_{i \in \mathcal{S}_{\mathrm{SIM}}}
\log
\frac{
\exp(s_{ii})
}{
\sum_{j \in \mathcal{S}_{\mathrm{SIM}}} \exp(s_{ij})
}.
\vspace*{-4pt}
\end{equation}
Here, $\mathcal{S}_{\mathrm{SIM}}$ denotes the supervised image positions in the batch. 
This objective is the embedding-space analogue of next-token prediction: each predicted visual embedding must identify its corresponding next embedding from a batch-level set of candidates. 
It also avoids the mean-prediction behavior that can arise with regression losses on approximately zero-mean patch embeddings.

\textbf{Cosine alternative.}
We also support a NEPA-style cosine-similarity objective:
\vspace*{-4pt}
\begin{equation}
\mathcal{L}_{\mathrm{SIM}}^{\mathrm{cos}}
=
-\frac{1}{|\mathcal{S}_{\mathrm{SIM}}|}
\sum_{t \in \mathcal{S}_{\mathrm{SIM}}}
\frac{
\hat{z}_{t+1}^{\top} z_{t+1}
}{
\lVert \hat{z}_{t+1} \rVert
\,
\lVert z_{t+1} \rVert
}.
\vspace*{-4pt}
\end{equation}
This provides a direction-only objective, but it is not on the same cross-entropy scale as the language modeling loss. 
We therefore use InfoNCE as the main image objective unless otherwise specified.

\subsection{Supervision Regimes}
\label{sec:supervision_regime}

We consider two regimes for deciding which positions contribute to the loss.

\textbf{Full-sequence.}
The default regime supervises all eligible text and image positions in the sampled sequence format. 
This is closest to standard pretraining on mixed visual-textual sequences: every supervised position contributes a next-token or next-embedding prediction loss, while the causal attention mask ensures that each prediction depends only on earlier context.

\textbf{Completion-only.}
The completion-only regime supervises only the response block associated with the sampled format: the post-condition for forward simulation, the pre-condition for backward simulation, and the action text for language modeling. 
This is analogous to supervised fine-tuning, where the prompt is used as conditioning context but only the response contributes to the loss.

%% file: floats/model-and-data.tex
\begin{figure}[!t]
    \centering
    \begin{minipage}[t]{0.53\linewidth}
    \vspace{0pt}
        \centering
        \includegraphics[width=\linewidth]{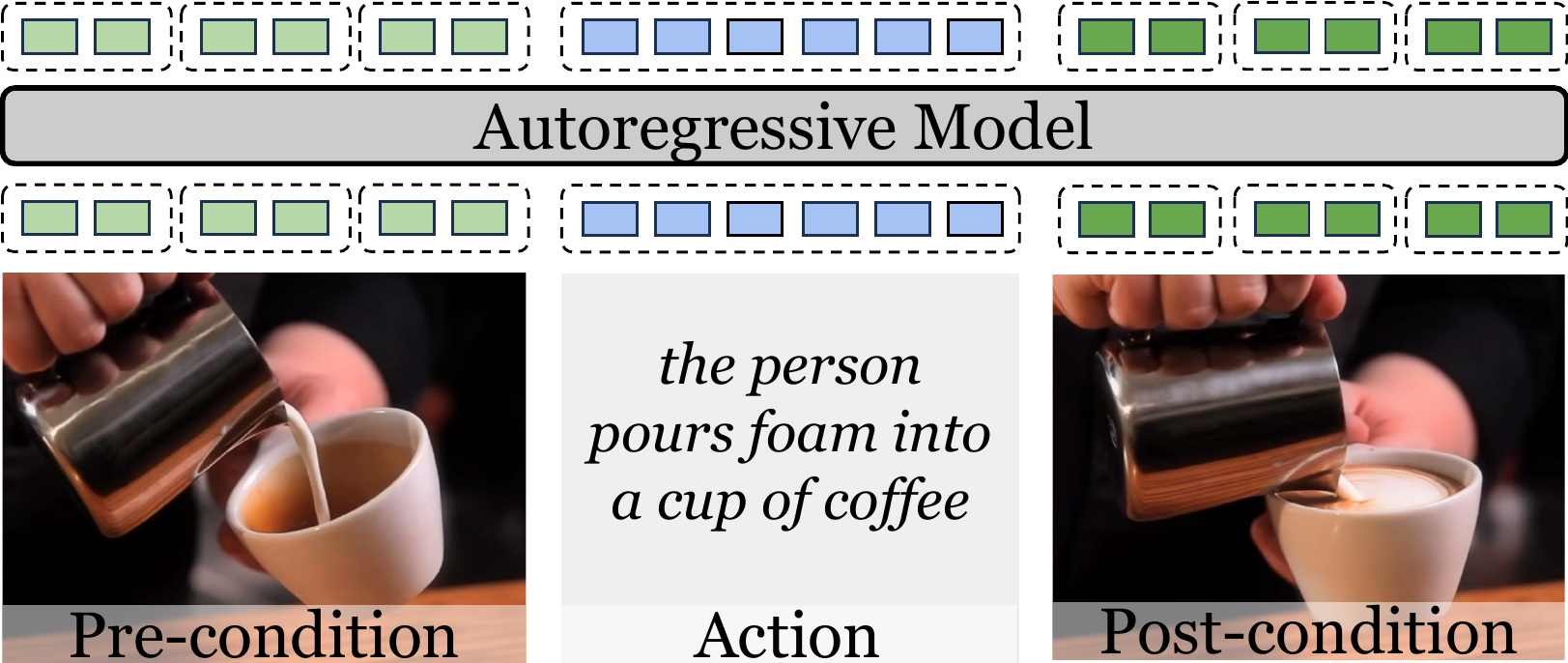}
        \vspace*{-15pt}
        \caption{
        Overview of our autoregressive world-language model, illustrated with
        \textbf{forward simulation}.
        }
        \label{fig:method}
    \end{minipage}
    \hfill
    \begin{minipage}[t]{0.43\linewidth}
        \centering
        \vspace{0pt}
        \captionof{table}{
            Data statistics for the real-video and
            world-model-distilled sources.
            Seen and unseen are curated subsets.
        }
        \vspace*{-5pt}
        \small
        \setlength{\tabcolsep}{2.5pt}

        \resizebox{\linewidth}{!}{%
            \begin{tabular}{llrrr}
                \toprule
                Source & Split & Triples & Verbs & Nouns \\
                \midrule
                Real-Video
                & Train 
                & 351{,}044 & 400 & 1{,}016 \\
                & Test
                & 3{,}999 & 80 & 407 \\
                \midrule
                \makecell[l]{World-Model\\Distilled}
                & Train 
                & 488{,}703 & 1{,}788 & 5{,}322 \\
                & Test (seen)
                & 47{,}332 & 97 & 1{,}220 \\
                & Test (unseen)
                & 12{,}097 & 120 & 794 \\
                \bottomrule
            \end{tabular}%
        }
        \vspace*{-3pt}
        
        \label{tab:dataset_stats}
    \end{minipage}

    \vspace*{-10pt}
\end{figure}

%% file: tables/sequence-format.tex
\begin{table}[t]
    \centering
    \caption{
        Sequence formats and their roles. $I_{\mathrm{pre}}$: pre-condition image. $I_{\mathrm{post}}$: post-condition image. $a$: action text.   Components are ordered from left to right. Underlining marks the response block under completion-only supervision, not the full-sequence loss mask.
        \vspace*{-5pt}
    }
    \label{tab:sequence_formats}

    \small
    \setlength{\tabcolsep}{5pt}
    \renewcommand{\arraystretch}{1.12}
    \begin{tabularx}{\linewidth}{
        @{}ll>{\raggedright\arraybackslash}X@{}
    }
        \toprule
        \textbf{Format} &
        \textbf{Sequence order} &
        \textbf{Description} \\
        \midrule

        $\mathrm{Sim}_{\mathrm{fwd}}$ &
        $I_{\mathrm{pre}},\; a,\; \underline{I_{\mathrm{post}}}$ &
        Predicts the resulting visual state from the initial state
        and action description.
        \\ \addlinespace[3pt]

        $\mathrm{Sim}_{\mathrm{bwd}}$ &
        $I_{\mathrm{post}},\; a,\; \underline{I_{\mathrm{pre}}}$ &
        Infers a compatible prior visual state from the observed
        result and action description.
        \\ \addlinespace[3pt]

        $\mathrm{LM}_{\mathrm{fwd}}$ &
        $I_{\mathrm{pre}},\; I_{\mathrm{post}},\; \underline{a}$ &
        Predicts the action text associated with the observed
        pre/post transition, corresponding to transition captioning.
        \\ \addlinespace[3pt]

        $\mathrm{LM}_{\mathrm{bwd}}$ &
        $I_{\mathrm{post}},\; I_{\mathrm{pre}},\; \underline{a}$ &
        Predicts the action text with the visual states reversed,
        testing whether gains arise from visual-state ordering
        rather than action-conditioned simulation.
        \\

        \bottomrule
    \end{tabularx}
    \vspace*{-5pt}
\end{table}

%% file: sec/04-setup.tex
\section{Experiment Setups}

We use two corpora shown in Table~\ref{tab:dataset_stats}.
The real videos are skewed toward a long tail of fine-grained manipulation, while the world-model-distilled corpus complements it with broader coverage of locomotion, environment change, viewpoint shift, and appearance change. 
We therefore use the former as a real-data sanity check under a narrower distribution, and the latter for the broader generalization study. 
We present more details in Appendices~\ref{app:dataset} and~\ref{app:experiment}.

\subsection{Real-Video Dataset}

We first construct a real-video dataset from 5 open video-understanding datasets: Assembly101~\citep{sener2022assembly101}, Ego4D~\citep{grauman2022ego4dworld3000hours}, EPIC-KITCHENS-100~\citep{Damen2018EPICKITCHENS,Damen2020RESCALING}, FineAction~\citep{liu2022fineaction}, and HoloAssist~\citep{HoloAssist2023}, which provide trimmed actions with labels and temporal boundaries.
For each source, we keep the training split and repurpose the validation split as our evaluation set. 

\textbf{Annotations.}
We convert each annotated segment into a triple by taking the start frame as the pre-condition and the end frame as the post-condition. 
Since the original action labels are short verb--noun tags with varying granularity (e.g., \textit{arrange (straighten, sort, distribute, align)} versus a bare \textit{cut}), we convert each clean verb--noun label into a short vision-conditioned sentence given the pre-condition and post-condition frames. 
We use Gemini-3-Flash\footnote{\url{https://ai.google.dev/gemini-api/docs/models/gemini-3-flash-preview}} for the best quality and throughput.
This step standardizes the action text and serves as a consistency filter, as examples whose generated sentence fails to preserve the cleaned verb or noun are corrected or removed.

\textbf{Test splits.}
Following~\citet{levin2008constraint}, we assign verb lemmas to \textit{manner} and \textit{result} classes using a manually curated verb inventory.
We then curate the test set by sampling 40 verbs from each class, with up to 50 examples per verb across available nouns. This subset allows us to test whether models rely more strongly on visual post-condition cues for result verbs. 



\subsection{World-Model-Distilled Dataset}

While real videos provide real temporal observations, their action coverage is source-limited, and designing controlled experiments is very hard.
To allow flexible ablations and broaden the range of action-conditioned transitions, we construct a complementary corpus from MSCOCO 2017 seed images~\citep{lin2014microsoft} and a visual generative model as the world model.

\textbf{Annotations.}
For each image, we use GPT-5.4-nano\footnote{\url{https://developers.openai.com/api/docs/models/gpt-5.4-nano}} to produce 4-5 diverse, physically and temporally plausible action candidates drawn from five action types: \textit{object interaction}, \textit{body motion}, \textit{scene transition}, \textit{state change}, \textit{new element}.
Each action candidate is converted into an editing instruction for FLUX.2 [klein]-9B~\citep{flux-2-2025}, which generates the post-condition image and forms a triple with the source image and action text.
We use GPT-5.5\footnote{\url{https://developers.openai.com/api/docs/models/gpt-5.5}} to filter inconsistent triples, followed by manual quality verification.
Generated triples are kept in the same split as their source images, with training-split triples used for training and test-split triples used for evaluation.

\textbf{Test splits.}
We report results on two test splits. 
The \(\mathrm{test\mbox{-}seen}\) split contains verb--object combinations observed during training. 
The \(\mathrm{test\mbox{-}unseen}\) split contains held-out verb–object combinations whose constituent verb and object lemmas each appear in other training combinations, allowing us to evaluate compositional generalization to unseen pairs.
Verb and object positions are annotated with a hybrid pipeline based on suffix matching, WordNet lemmatization, and a small hand-curated override table for irregular forms and plural-only nouns. 
Each test example contains a verb index, an object index, and canonical lemma identifiers, allowing surprisal to be aggregated by lemma. 

\subsection{Evaluation Protocol and Metrics}

We evaluate word learning by measuring the \textit{predictability} of action words and their participants from visual context. 
We use the forward language modeling format: the two visual states are provided before the action sentence, and the model is evaluated by next-token prediction at the annotated positions. 
When available, target words are mapped to canonical lemma identifiers, so that results can be aggregated by concept rather than only by surface form.

\textbf{Word surprisal.}
Let \(w_k\) denote a target word in the action sentence $\mathcal{S}$, such as an annotated verb or object token. 
Given the visual context \(c = (\mathrm{pre\mbox{-}condition}, \mathrm{post\mbox{-}condition})\) and the preceding action prefix \(a_{<k}\), we define target-word surprisal as
\vspace*{-3pt}
\begin{equation}
s_k(\mathcal{S}) = -\log_2 P(w_k \mid c, a_{<k}).
\vspace*{-3pt}
\label{eq:surprisal}
\end{equation}
When multiple target words are annotated for the same example, we report their average surprisal:
\vspace*{-3pt}
\begin{equation}
s_{\mathrm{target}}(\mathcal{S})
=
\frac{1}{|\mathcal{K}|}
\sum_{k \in \mathcal{K}} s_k(\mathcal{S}),
\vspace*{-3pt}
\end{equation}
where \(\mathcal{K}\) is the set of valid target positions. 

\textbf{Learning curves and neural age of acquisition.}
Following \citet{ma2025babysit}, we summarize acquisition dynamics with both fitted learning curves and neural age of acquisition (nAoA). 
For each target word, we evaluate surprisal on the test set across training checkpoints and fit a learning curve over log training steps, capturing how predictability improves over the course of training. 
From the fitted curve, we compute nAoA as the training step at which the target crosses a specified acquisition threshold; lower nAoA indicates earlier acquisition. 
We report nAoA at the standard 0.50 threshold and, for robustness, also average nAoA over thresholds from 0.50 to 0.95. 

%% file: sec/05-findings.tex
\section{Main Findings}
\label{sec:findings}

\input{tables/surprisal}

Table~\ref{tab:results} summarizes the main surprisal results at step $10{,}000$ on the world-model-distilled dataset, and Table~\ref{tab:video-results} reports the corresponding evaluation on real trimmed action videos. 
Figure~\ref{fig:aoa-infonce-fullseq} and Table~\ref{tab:aoa-fullseq-infonce} complement the final-step results with learning-curve fits and nAoA results. 
Figure~\ref{fig:simact-example-all} shows qualitative examples.
The central pattern is consistent across datasets: simulation is most useful as an auxiliary objective to language modeling, improving action-word and verb--noun predictability while leaving noun learning comparatively less affected.

\input{tables/surprisal-real-vid}
\input{tables/aoa}

\textbf{Simulation improves grounding when text supervision is sufficient.}
The best-performing configurations allocate half the training budget to forward language modeling and half to simulation.
In Table~\ref{tab:results}, these mixtures consistently improve verb and pair surprisal over the language-only baseline, with the balanced bidirectional mix performing best. 
By contrast, the symmetric mixture with only one third language modeling is weaker, both in final surprisal and in acquisition speed. 
This suggests that simulation supplies useful visual structure when the model has enough direct exposure to action prediction to convert that structure into lexical gains.
Similar improvements are observed under alternative visual losses and supervision regimes.
\input{figures/results}

\textbf{Backward simulation contributes more than forward simulation.}
Although balanced bidirectional simulation gives the strongest final surprisal in Table~\ref{tab:results}, backward simulation gives the clearest improvement in verb acquisition speed in Table~\ref{tab:aoa-fullseq-infonce}. 
This suggests that inverse prediction is a more diagnostic auxiliary task for action meaning~\citep{baker2009action}: recovering the pre-condition from the post-condition and action requires representing what the action changed, whereas forward prediction can partly exploit visual continuity between the two states.

\textbf{The pattern transfers to real action videos.}
The real-video results in Table~\ref{tab:video-results} show that the same principle extends beyond the controlled before/after setting. 
Despite the noisier visual input, adding simulation improves final lexical predictability, and the balanced bidirectional mixture performs best across the reported manner/result verb and noun columns. 
This strengthens the main claim: visual simulation is not merely exploiting artifacts of a controlled transition dataset, but provides a useful auxiliary signal for grounding action language in naturalistic visual observations.

\textbf{Simulation accelerates verb learning, but not noun learning.}
Table~\ref{tab:aoa-fullseq-infonce} shows that simulation advances verb nAoA by $0.13$--$0.19$ log$_{10}$ steps, while noun nAoA changes by only $0.02$--$0.06$.
Figure~\ref{fig:aoa-infonce-fullseq} shows the same pattern visually.
Nouns can often be grounded by direct reference to visible entities, whereas verbs depend on the relation between pre-condition and post-condition. 

\input{tables/manner-result}

\textbf{Result verbs benefit more from simulation than manner verbs.}
Table~\ref{tab:manner-result-fullseq-infonce} shows the clearest difference in acquisition speed. Across the three 50\%-LM simulation settings, result-verb nAoA has $\Delta=-0.15,-0.09,-0.14$, compared with $\Delta=+0.01,+0.02,-0.02$ for manner verbs.
Final surprisal improves for both classes, with balanced bidirectional simulation also giving a slightly larger gain for result verbs ($-0.63$ vs.\ $-0.58$).
This pattern is consistent with the semantic distinction between the two classes. 
Result verbs are closely tied to visible changes of state, which are directly exposed by before/after prediction. 
Manner verbs depend more on how an action unfolds over time, so a static transition signal is useful but incomplete. 


\input{figures/examples/combo}


%% file: tables/surprisal.tex
\begin{table*}[!t]
\centering
\caption{
    On the world-model-distilled dataset, surprisal results at training step 10,000, pooled over 3 random seeds. The four \texttt{Supervision} columns give the sampling probability of each base format: \texttt{LM$_\text{bwd}$} (\textit{postcon}$+$\textit{precon}$\rightarrow$\textit{action}), \texttt{LM$_\text{fwd}$} (\textit{precon}$+$\textit{postcon}$\rightarrow$\textit{action}), \texttt{Sim$_\text{bwd}$} (\textit{postcon}$+$\textit{action}$\rightarrow$\textit{precon}), and \texttt{Sim$_\text{fwd}$} (\textit{precon}$+$\textit{action}$\rightarrow$\textit{postcon}). The Generalization Gap is the aggregate pair-surprisal gap, Unseen$-$Seen. 
    \vspace*{-8pt}
}
    \resizebox{\textwidth}{!}{
    \begingroup
    \setlength{\tabcolsep}{2.5pt}
    \renewcommand{\arraystretch}{0.9}
    \begin{tabular}{ccccccccccccc}
    \toprule
    \multicolumn{4}{c}{Supervision} & \multicolumn{2}{c}{Loss Configuration} & \multicolumn{2}{c}{Verb Surprisal $\downarrow$} & \multicolumn{2}{c}{Noun Surprisal $\downarrow$} & \multicolumn{2}{c}{Pair Surprisal $\downarrow$} & Gen. Gap $\downarrow$ \\
    \cmidrule(r){1-4} \cmidrule(r){5-6} \cmidrule(r){7-8} \cmidrule(r){9-10} \cmidrule(r){11-12} \cmidrule(r){13-13}
    LM$_\text{bwd}$ & LM$_\text{fwd}$ & Sim$_\text{bwd}$ & Sim$_\text{fwd}$ & Applied To & Visual & Seen & Unseen & Seen & Unseen & Seen & Unseen & Pair (U$-$S) \\
    \midrule
    100\% & -- & -- & -- & \texttt{Full Seq} & \texttt{InfoNCE} & $9.06_{\pm.05}$ & $9.13_{\pm.04}$ & $2.50_{\pm.03}$ & $6.06_{\pm.04}$ & $5.78_{\pm.03}$ & $7.59_{\pm.03}$ & $+1.82_{\pm.04}$ \\
    50\% & 50\% & -- & -- & \texttt{Full Seq} & \texttt{InfoNCE} & $8.24_{\pm.06}$ & $8.60_{\pm.05}$ & $1.53_{\pm.03}$ & $4.68_{\pm.04}$ & $4.88_{\pm.03}$ & $6.64_{\pm.03}$ & $+1.76_{\pm.04}$ \\
    -- & 100\% & -- & -- & \texttt{Full Seq} & \texttt{InfoNCE} & $8.31_{\pm.06}$ & $8.53_{\pm.05}$ & $1.55_{\pm.03}$ & $4.78_{\pm.04}$ & $4.93_{\pm.03}$ & $6.65_{\pm.03}$ & $+1.72_{\pm.04}$ \\
    -- & 50\% & 50\% & -- & \texttt{Full Seq} & \texttt{InfoNCE} & $7.81_{\pm.05}$ & $8.15_{\pm.05}$ & $\sota{1.49}_{\pm.02}$ & $4.70_{\pm.04}$ & $4.65_{\pm.03}$ & $6.43_{\pm.03}$ & $+1.78_{\pm.04}$ \\
    -- & 50\% & -- & 50\% & \texttt{Full Seq} & \texttt{InfoNCE} & $8.17_{\pm.05}$ & $8.41_{\pm.05}$ & $1.66_{\pm.03}$ & $5.01_{\pm.04}$ & $4.92_{\pm.03}$ & $6.71_{\pm.03}$ & $+1.79_{\pm.04}$ \\
    -- & 50\% & 25\% & 25\% & \texttt{Full Seq} & \texttt{InfoNCE} & $\sota{7.66}_{\pm.05}$ & $\sota{8.05}_{\pm.05}$ & $1.56_{\pm.02}$ & $\sota{4.58}_{\pm.04}$ & $\sota{4.61}_{\pm.03}$ & $\sota{6.32}_{\pm.03}$ & $\sota{+1.71}_{\pm.04}$ \\
    -- & 33\% & 33\% & 33\% & \texttt{Full Seq} & \texttt{InfoNCE} & $8.28_{\pm.06}$ & $8.61_{\pm.05}$ & $1.58_{\pm.03}$ & $4.83_{\pm.04}$ & $4.93_{\pm.03}$ & $6.72_{\pm.03}$ & $+1.79_{\pm.04}$ \\
    \midrule
    100\% & -- & -- & -- & \texttt{Full Seq} & \texttt{Cosine} & $9.60_{\pm.06}$ & $9.30_{\pm.05}$ & $2.42_{\pm.03}$ & $6.06_{\pm.04}$ & $6.01_{\pm.03}$ & $7.68_{\pm.03}$ & $\sota{+1.67}_{\pm.04}$ \\
    50\% & 50\% & -- & -- & \texttt{Full Seq} & \texttt{Cosine} & $7.85_{\pm.05}$ & $8.31_{\pm.04}$ & $1.62_{\pm.02}$ & $5.20_{\pm.04}$ & $4.73_{\pm.03}$ & $6.76_{\pm.03}$ & $+2.02_{\pm.04}$ \\
    -- & 100\% & -- & -- & \texttt{Full Seq} & \texttt{Cosine} & $8.18_{\pm.05}$ & $8.59_{\pm.05}$ & $1.66_{\pm.03}$ & $5.30_{\pm.04}$ & $4.92_{\pm.03}$ & $6.94_{\pm.03}$ & $+2.02_{\pm.04}$ \\
    -- & 50\% & 50\% & -- & \texttt{Full Seq} & \texttt{Cosine} & $\sota{7.30}_{\pm.05}$ & $\sota{7.89}_{\pm.04}$ & $\sota{1.58}_{\pm.02}$ & $5.16_{\pm.04}$ & $\sota{4.44}_{\pm.03}$ & $\sota{6.52}_{\pm.03}$ & $+2.08_{\pm.04}$ \\
    -- & 50\% & -- & 50\% & \texttt{Full Seq} & \texttt{Cosine} & $8.23_{\pm.05}$ & $8.49_{\pm.04}$ & $1.76_{\pm.03}$ & $5.39_{\pm.04}$ & $5.00_{\pm.03}$ & $6.94_{\pm.03}$ & $+1.94_{\pm.04}$ \\
    -- & 50\% & 25\% & 25\% & \texttt{Full Seq} & \texttt{Cosine} & $8.32_{\pm.05}$ & $8.57_{\pm.05}$ & $1.67_{\pm.03}$ & $\sota{5.11}_{\pm.04}$ & $5.00_{\pm.03}$ & $6.84_{\pm.03}$ & $+1.85_{\pm.04}$ \\
    -- & 33\% & 33\% & 33\% & \texttt{Full Seq} & \texttt{Cosine} & $7.68_{\pm.05}$ & $8.16_{\pm.04}$ & $1.69_{\pm.03}$ & $5.18_{\pm.04}$ & $4.68_{\pm.03}$ & $6.67_{\pm.03}$ & $+1.98_{\pm.04}$ \\
    \midrule
    100\% & -- & -- & -- & \texttt{Completion} & -- & $9.58_{\pm.06}$ & $9.39_{\pm.05}$ & $2.35_{\pm.03}$ & $6.12_{\pm.04}$ & $5.97_{\pm.03}$ & $7.76_{\pm.03}$ & $\sota{+1.79}_{\pm.04}$ \\
    50\% & 50\% & -- & -- & \texttt{Completion} & -- & $9.13_{\pm.06}$ & $9.06_{\pm.05}$ & $1.78_{\pm.03}$ & $5.65_{\pm.04}$ & $5.46_{\pm.03}$ & $7.36_{\pm.03}$ & $+1.90_{\pm.04}$ \\
    -- & 100\% & -- & -- & \texttt{Completion} & -- & $8.25_{\pm.05}$ & $8.58_{\pm.05}$ & $1.68_{\pm.02}$ & $5.32_{\pm.04}$ & $4.97_{\pm.03}$ & $6.95_{\pm.03}$ & $+1.98_{\pm.04}$ \\
    -- & 50\% & 50\% & -- & \texttt{Completion} & \texttt{InfoNCE} & $7.76_{\pm.05}$ & $\sota{8.31}_{\pm.05}$ & $1.69_{\pm.02}$ & $4.96_{\pm.04}$ & $4.72_{\pm.03}$ & $6.64_{\pm.03}$ & $+1.91_{\pm.04}$ \\
    -- & 50\% & -- & 50\% & \texttt{Completion} & \texttt{InfoNCE} & $\sota{7.75}_{\pm.05}$ & $8.36_{\pm.04}$ & $1.69_{\pm.02}$ & $4.94_{\pm.04}$ & $\sota{4.72}_{\pm.03}$ & $6.65_{\pm.03}$ & $+1.93_{\pm.04}$ \\
    -- & 50\% & 25\% & 25\% & \texttt{Completion} & \texttt{InfoNCE} & $7.82_{\pm.05}$ & $8.33_{\pm.05}$ & $\sota{1.65}_{\pm.02}$ & $\sota{4.82}_{\pm.03}$ & $4.73_{\pm.03}$ & $\sota{6.58}_{\pm.03}$ & $+1.84_{\pm.04}$ \\
    -- & 33\% & 33\% & 33\% & \texttt{Completion} & \texttt{InfoNCE} & $8.59_{\pm.05}$ & $8.94_{\pm.05}$ & $1.87_{\pm.03}$ & $5.39_{\pm.04}$ & $5.23_{\pm.03}$ & $7.17_{\pm.03}$ & $+1.94_{\pm.04}$ \\
    \bottomrule
    \end{tabular}
    \endgroup}
\vspace*{-9pt}

\label{tab:results}
\end{table*}

%% file: tables/surprisal-real-vid.tex
\begin{table}[!t]
\centering
\caption{
    On the real-video test set, Full-sequence $+$ InfoNCE loss, pooled over 3 random seeds.
    \vspace*{-8pt}
}
    \resizebox{0.52\textwidth}{!}{
    \begingroup
    \setlength{\tabcolsep}{2pt}
    \renewcommand{\arraystretch}{0.9}
    \begin{tabular}{ccccccc}
    \toprule
    \multicolumn{3}{c}{Supervision} & \multicolumn{2}{c}{Manner Surprisal $\downarrow$} & \multicolumn{2}{c}{Result Surprisal $\downarrow$} \\
    \cmidrule(r){1-3} \cmidrule(r){4-5} \cmidrule(r){6-7}
    LM$_\text{fwd}$ & Sim$_\text{bwd}$ & Sim$_\text{fwd}$ & Verb & Noun & Verb & Noun \\
    \midrule
    100\% & -- & -- & $3.30_{\pm .02}$ & $2.45_{\pm .02}$ & $3.94_{\pm .02}$ & $3.22_{\pm .04}$ \\
    50\% & 50\% & -- & $3.21_{\pm .00}$ & $2.34_{\pm .02}$ & $3.81_{\pm .00}$ & $3.05_{\pm .01}$ \\
    50\% & -- & 50\% & $3.27_{\pm .04}$ & $2.40_{\pm .01}$ & $3.81_{\pm .01}$ & $3.08_{\pm .01}$ \\
    50\% & 25\% & 25\% & $\sota{3.15}_{\pm .02}$ & $\sota{2.28}_{\pm .02}$ & $\sota{3.66}_{\pm .01}$ & $\sota{2.97}_{\pm .01}$ \\
    \bottomrule
    \end{tabular}
    \endgroup}
\vspace*{-5pt}

\label{tab:video-results}
\end{table}

%% file: tables/aoa.tex
\begin{table*}[!t]
\centering
\caption{nAoA values for Full-sequence + InfoNCE from a single double-sigmoid fit per (mix, metric, split) to the 3-seed-pooled trajectory (step $\le$ 10,000). 
\texttt{@[.5]}: $\log_{10}(\text{step})$ at which the fitted curve crosses the midpoint of the observed range. \texttt{@[.5:.95]}: average of the same crossing across progress thresholds $\{0.50, 0.55, \dots, 0.95\}$. $\Delta$ denotes improvement over the 100\% LM baseline
(current setting $-$ baseline), so more negative values indicate larger gains. \texttt{All} pools seen$+$unseen combos into one trajectory before fitting.
\vspace*{-8pt}
}
\resizebox{\textwidth}{!}{
\setlength{\tabcolsep}{2pt}
\begin{tabular}{ccccccccccccccc}
\toprule
\multicolumn{3}{c}{Supervision} & \multicolumn{6}{c}{Verb nAoA $\downarrow$} & \multicolumn{6}{c}{Noun nAoA $\downarrow$} \\
\cmidrule(r){1-3} \cmidrule(r){4-9} \cmidrule(r){10-15}
  &  & & \multicolumn{2}{c}{Seen} & \multicolumn{2}{c}{Unseen} & \multicolumn{2}{c}{All} & \multicolumn{2}{c}{Seen} & \multicolumn{2}{c}{Unseen} & \multicolumn{2}{c}{All} \\
\cmidrule(r){4-5} \cmidrule(r){6-7} \cmidrule(r){8-9} \cmidrule(r){10-11} \cmidrule(r){12-13} \cmidrule(r){14-15}
LM$_\text{fwd}$ & Sim$_\text{bwd}$ & Sim$_\text{fwd}$ & @[.5] & @[.5:.95] & @[.5] & @[.5:.95] & @[.5] & @[.5:.95] & @[.5] & @[.5:.95] & @[.5] & @[.5:.95] & @[.5] & @[.5:.95] \\
\midrule
100\% & -- & -- & $2.73$ & $3.13$ & $2.66$ & $3.11$ & $2.70$ & $3.12$ & $2.61$ & $2.96$ & $2.81$ & $3.17$ & $2.69$ & $3.05$ \\
50\% & 50\% & -- & $\sota{2.54}$ & $\sota{2.98}$ & $\sota{2.53}$ & $\sota{2.98}$ & $\sota{2.53}$ & $\sota{2.97}$ & $2.61$ & $2.96$ & $\sota{2.75}$ & $\sota{3.13}$ & $2.67$ & $3.03$ \\
50\% & -- & 50\% & $2.62$ & $3.04$ & $2.57$ & $3.01$ & $2.59$ & $3.03$ & $2.62$ & $2.97$ & $2.77$ & $3.16$ & $2.68$ & $3.04$ \\
50\% & 25\% & 25\% & $2.58$ & $3.00$ & $2.54$ & $\sota{2.98}$ & $2.56$ & $2.99$ & $\sota{2.58}$ & $\sota{2.94}$ & $\sota{2.75}$ & $3.14$ & $\sota{2.64}$ & $\sota{3.02}$ \\
33\% & 33\% & 33\% & $2.77$ & $3.23$ & $2.63$ & $3.15$ & $2.70$ & $3.19$ & $\sota{2.58}$ & $2.96$ & $2.77$ & $3.17$ & $2.65$ & $3.05$ \\
\midrule
\multicolumn{3}{c}{Best $\Delta$ vs.\ LM}
& $\mathllap{-}0.19$ & $\mathllap{-}0.15$
& $\mathllap{-}0.13$ & $\mathllap{-}0.13$
& $\mathllap{-}0.17$ & $\mathllap{-}0.15$
& $\mathllap{-}0.03$ & $\mathllap{-}0.02$
& $\mathllap{-}0.06$ & $\mathllap{-}0.04$
& $\mathllap{-}0.05$ & $\mathllap{-}0.03$ \\
\bottomrule
\end{tabular}}
\vspace*{-15pt}

\label{tab:aoa-fullseq-infonce}
\end{table*}

%% file: figures/results.tex
\begin{figure*}[!t]
    \vspace{15pt}
    \centering
    \begin{subfigure}[t]{.31\textwidth}
        \centering
        \includegraphics[width=1.05\linewidth]{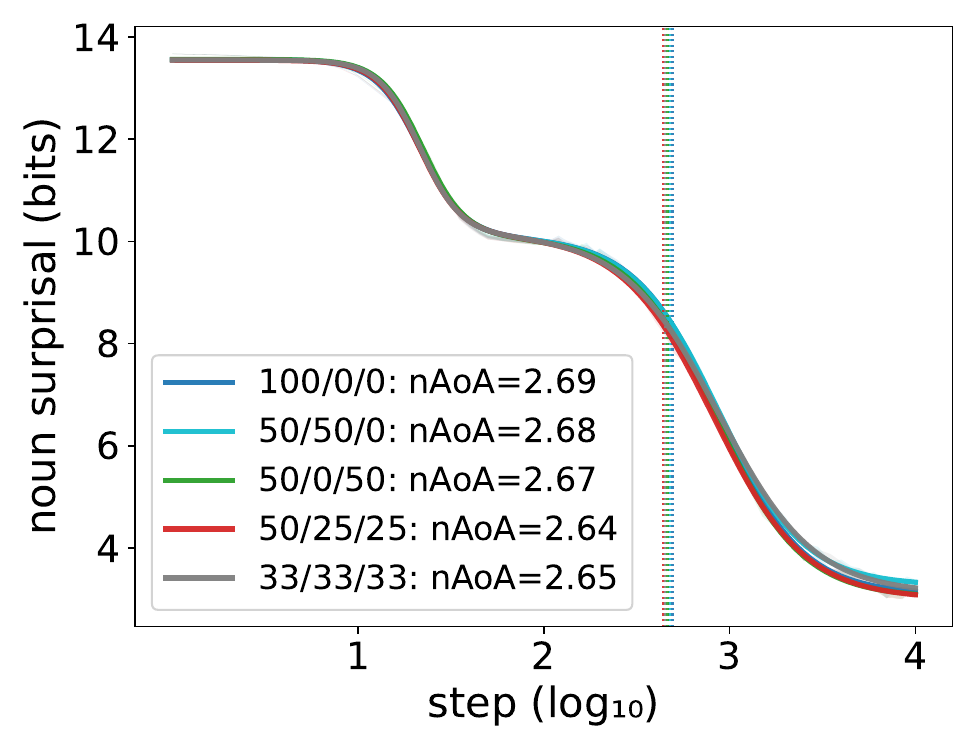}
        \vspace*{-20pt}
        \caption{Noun surprisal.}
        \label{fig:noun-infonce-fullseq}
    \end{subfigure}
    ~
    \begin{subfigure}[t]{.31\textwidth}
        \centering
        \includegraphics[width=1.05\linewidth]{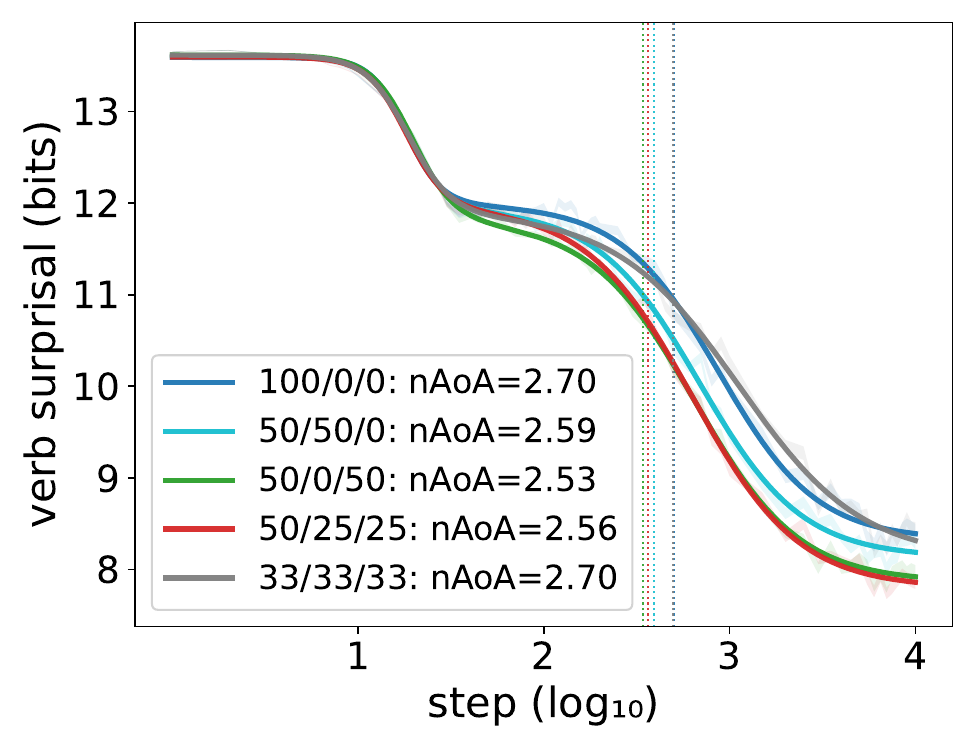}
        \vspace*{-20pt}
        \caption{Verb surprisal.}
        \label{fig:verb-infonce-fullseq}
    \end{subfigure}
    ~
    \begin{subfigure}[t]{.31\textwidth}
        \centering
        \includegraphics[width=1.05\linewidth]{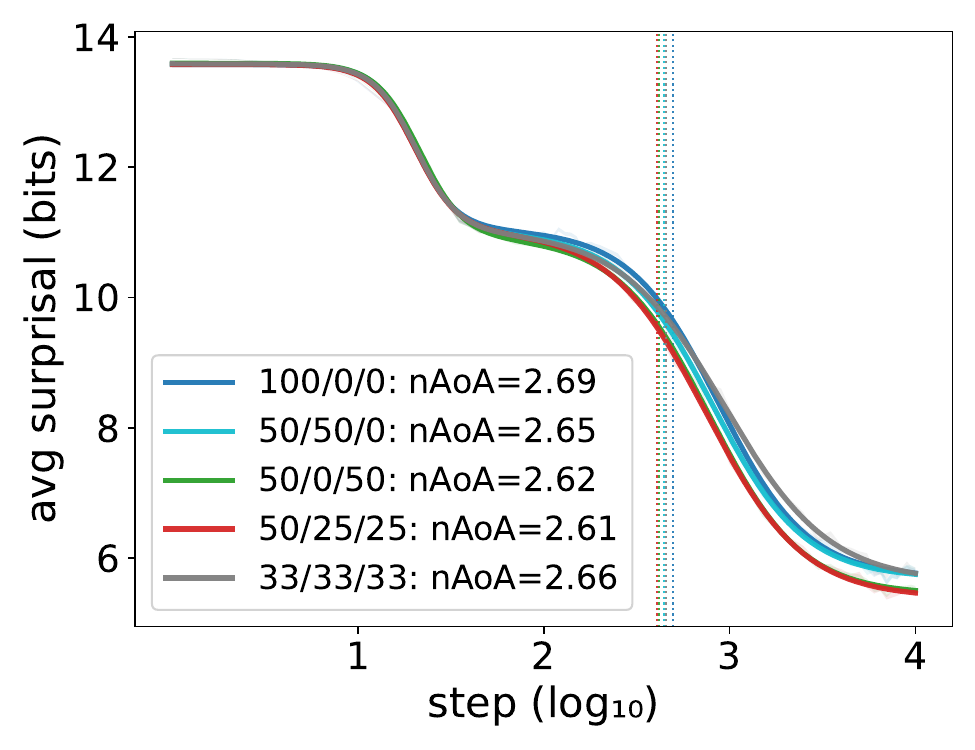}
        \vspace*{-20pt}
        \caption{Pair surprisal.}
        \label{fig:combined-infonce-fullseq}
    \end{subfigure}
    \vspace*{-5pt}
    \caption{
        Fitted learning curves for the Full-sequence $+$ InfoNCE loss configuration, comparing five supervision mixes (legend: \texttt{Fwd-LM/Fwd-Sim/Bwd-Sim} sampling \%). 
        Each curve is a double-sigmoid fit to the mean surprisal trajectory pooled over 3 random seeds and seen$+$unseen test combos. Vertical dotted lines mark $\texttt{nAoA@0.50}$, the $\log_{10}(\text{step})$ at which each curve crosses the midpoint of its observed range.
        \vspace*{-15pt}
    }
    \label{fig:aoa-infonce-fullseq}
\end{figure*}

%% file: tables/manner-result.tex
\begin{table}[!t]
\centering
\hspace*{-10pt}

\caption{Manner vs result verb classes, Full-sequence $+$ InfoNCE loss (step $\le$ 10,000). \textbf{Final verb surprisal}: per-verb mean (pooled over combos and 3 seeds) at the final step, then averaged across verbs in the class. \textbf{nAoA@[.5:.95]}: average crossing across progress thresholds $\{0.50, 0.55, \dots, 0.95\}$, averaged across verbs that fit successfully. Learning curves are fit using approximately 57K examples from the test set. \texttt{Supervision} columns give the sampling probability of each base format. $n_\text{manner}=34$, $n_\text{result}=54$ for surprisal; $n_\text{manner}=32$, $n_\text{result}=50$ for nAoA. $\Delta$ denotes the difference from the 100\% forward-LM baseline (current setting $-$ baseline), so more negative values indicate larger gains. 
}

\resizebox{0.65\textwidth}{!}{%
\setlength{\tabcolsep}{2pt}
\begin{tabular}{ccccccccccc}
\toprule
\multicolumn{3}{c}{Supervision}
& \multicolumn{4}{c}{Surprisal $\downarrow$}
& \multicolumn{4}{c}{nAoA@$[$.5:.95$]$ $\downarrow$} \\
\cmidrule(r){1-3}
\cmidrule(r){4-7}
\cmidrule(r){8-11}
LM$_\text{fwd}$ & Sim$_\text{bwd}$ & Sim$_\text{fwd}$ & Man. & $\Delta$ & Res. & $\Delta$ & Man. & $\Delta$ & Res. & $\Delta$ \\
\cmidrule(r){1-3}
\cmidrule(r){4-7}
\cmidrule(r){8-11}
100\% & -- & -- & $8.14$ & -- & $8.84$ & -- & $2.99$ & -- & $2.89$ & --\\
50\% & 50\% & -- 
& $7.68$ & $-0.46$ 
& $8.41$ & $-0.43$ 
& $3.00$ & $+0.01$ 
& $\sota{2.74}$ & $\sota{-0.15}$ \\
50\% & -- & 50\% 
& $7.99$ & $-0.15$ 
& $8.72$ & $-0.12$
& $3.01$ & $+0.02$
& $2.80$ & $-0.09$ \\
50\% & 25\% & 25\% 
& $\sota{7.56}$ & $\sota{-0.58}$
& $\sota{8.21}$ & $\sota{-0.63}$
& $\sota{2.97}$ & $\sota{-0.02}$
& $2.75$ & $-0.14$ \\
33\% & 33\% & 33\% 
& $8.29$ & $+0.15$
& $8.79$ & $-0.05$
& $2.99$ & $0.00$
& $2.95$ & $+0.06$ \\
\bottomrule
\end{tabular}}

\vspace*{-10pt}
\label{tab:manner-result-fullseq-infonce}
\end{table}

%% file: figures/examples/combo.tex
\begin{figure*}[h!]
    \centering
    \captionsetup[subfigure]{font=small,labelfont=bf,skip=2pt}
    \simactheader
    \simactrowcustom
        {\simactcelltune{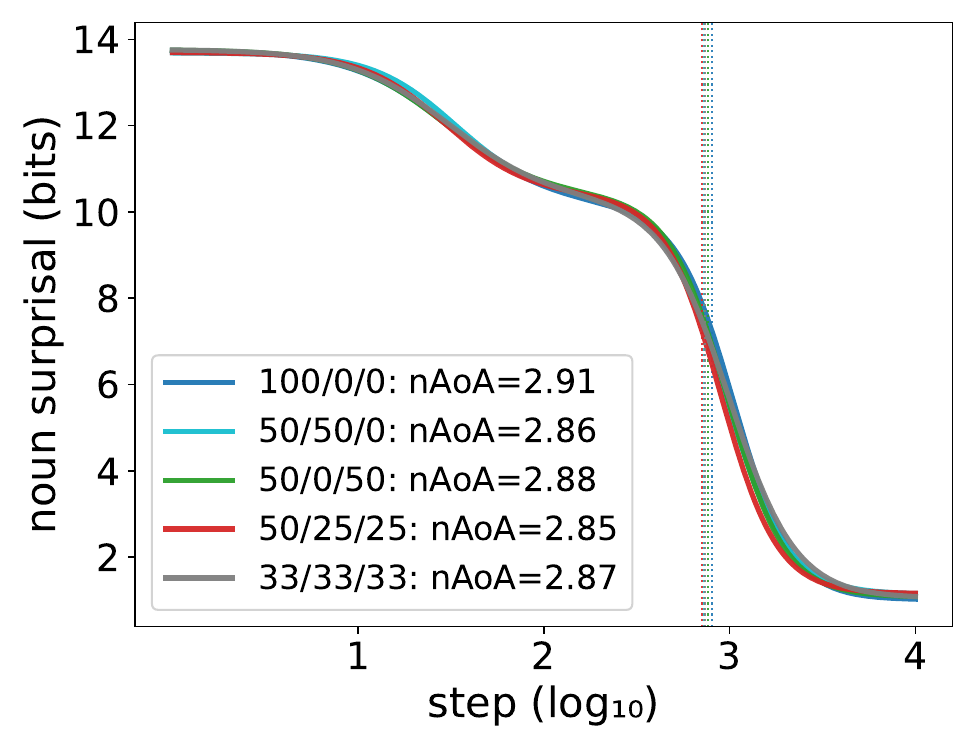}{1.0}{1.0}{0pt}{0 0 0 0}{\simactcellwL}}
        {\simactcelltune{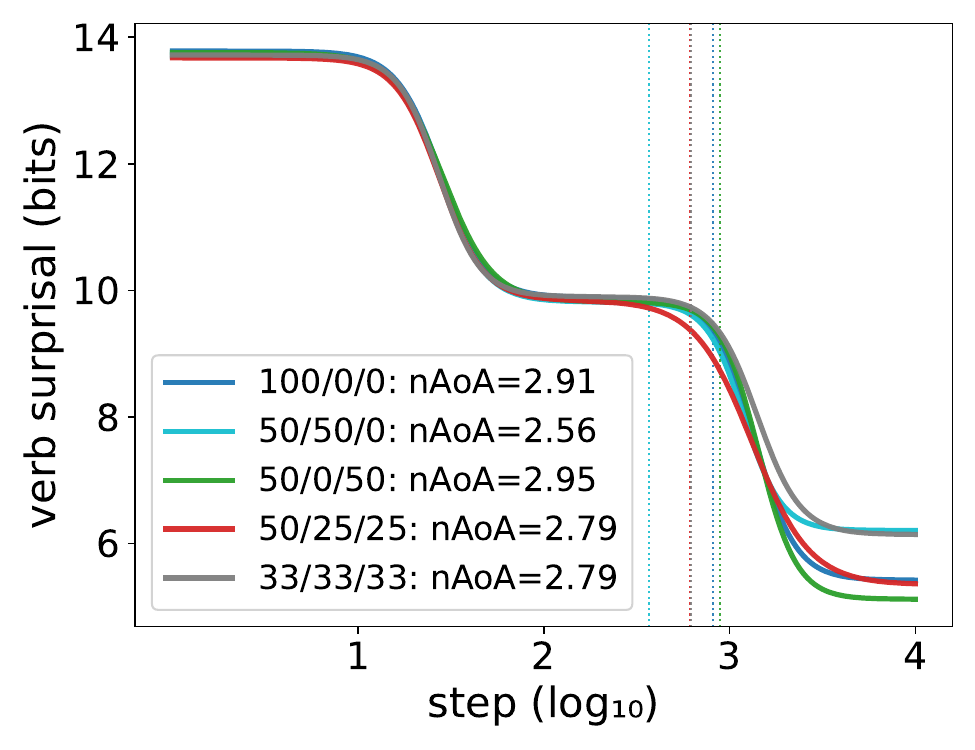}{1.0}{1.0}{0pt}{0 0 0 0}{\simactcellwL}}
        {\simactcelltune{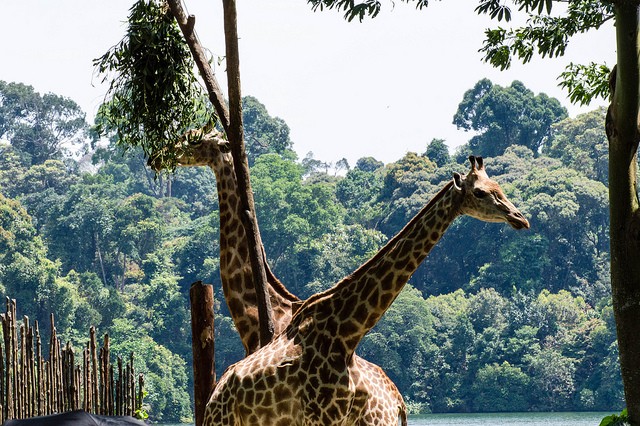}{1.00}{0.9}{7pt}{0 0 0 0}{\simactcellwR}}
        {\simactcelltune{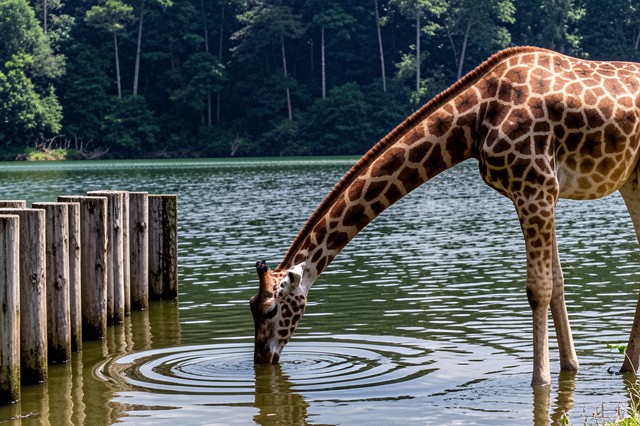}{1.00}{0.9}{7pt}{0 0 0 0}{\simactcellwR}}
        {Seen pair: \textit{lower head}.}
        {fig:simact-example-lower-head}
    \simactrowcustom
        {\simactcelltune{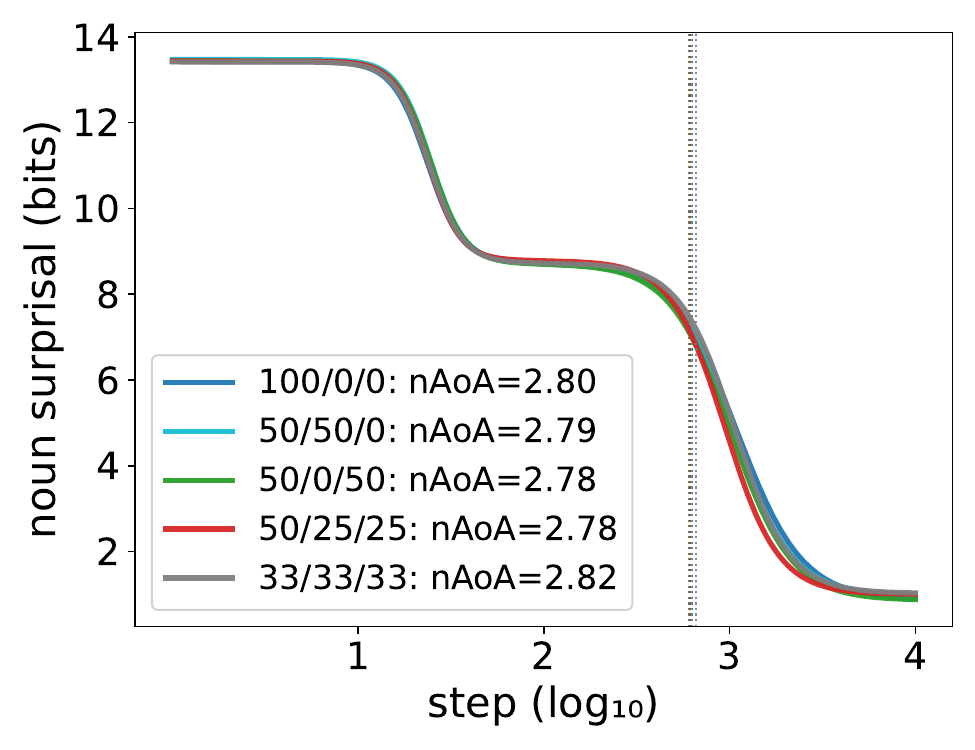}{1.0}{1.00}{0pt}{0 0 0 0}{\simactcellwL}}
        {\simactcelltune{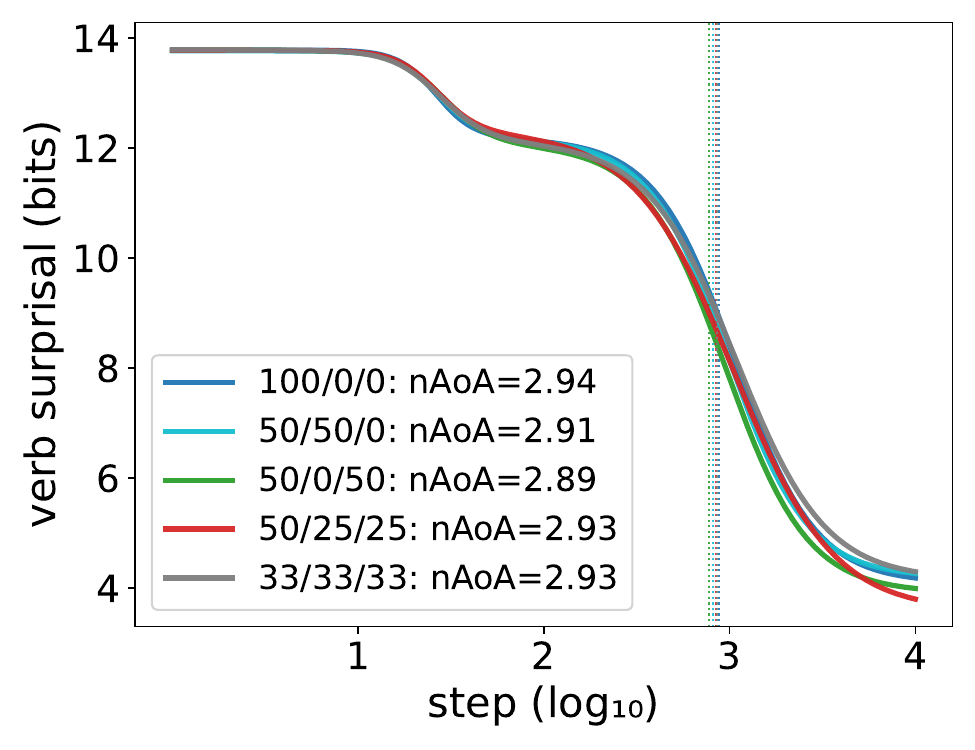}{1.0}{1.00}{0pt}{0 0 0 0}{\simactcellwL}}
        {\simactcelltune{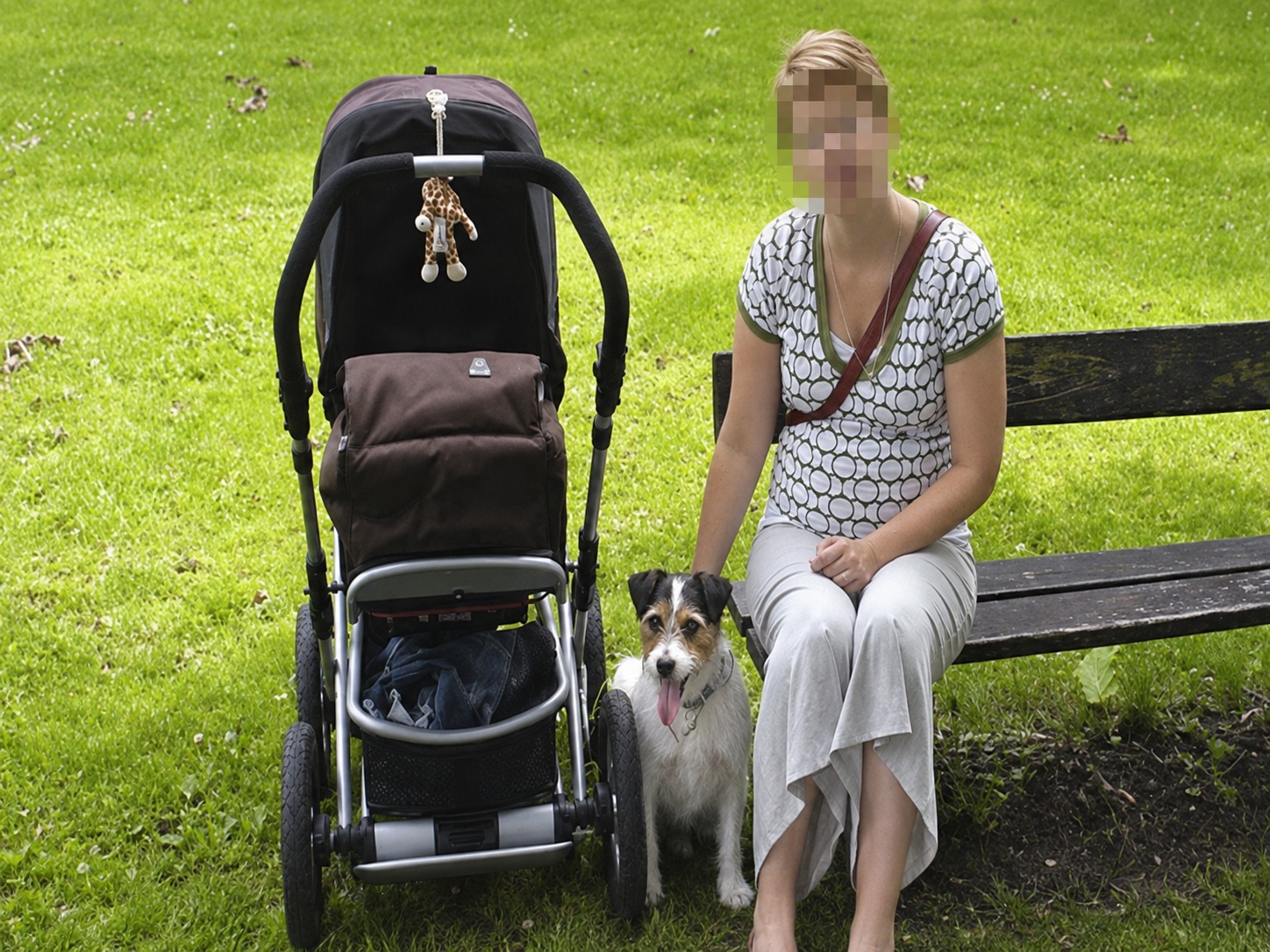}{0.95}{0.95}{5pt}{0 0 0 0}{\simactcellwR}}
        {\simactcelltune{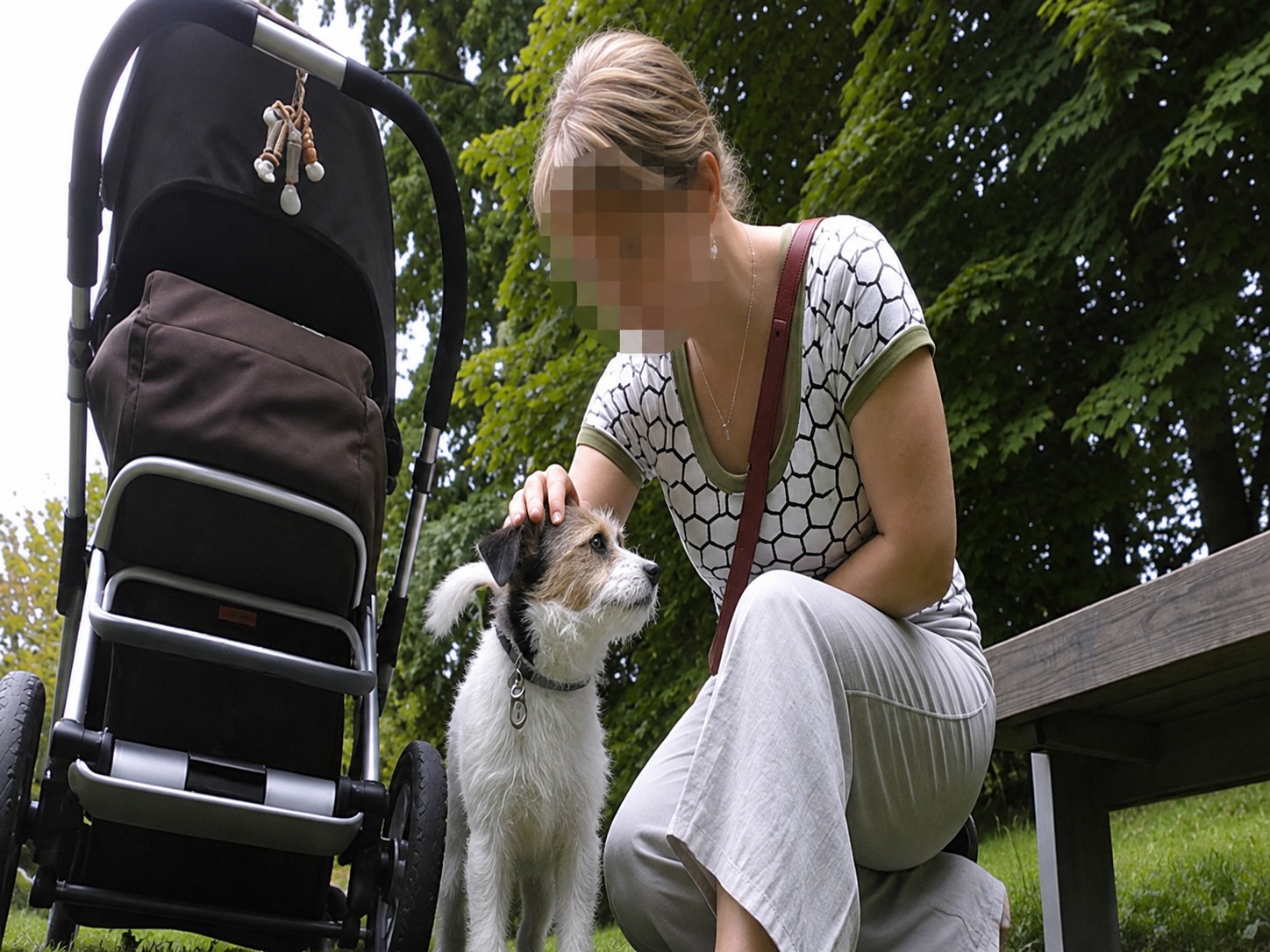}{0.95}{0.95}{5pt}{0 0 0 0}{\simactcellwR}}
        {Seen pair: \textit{pet dog}.}
        {fig:simact-example-pet-dog}
    \simactrowcustom
        {\simactcelltune{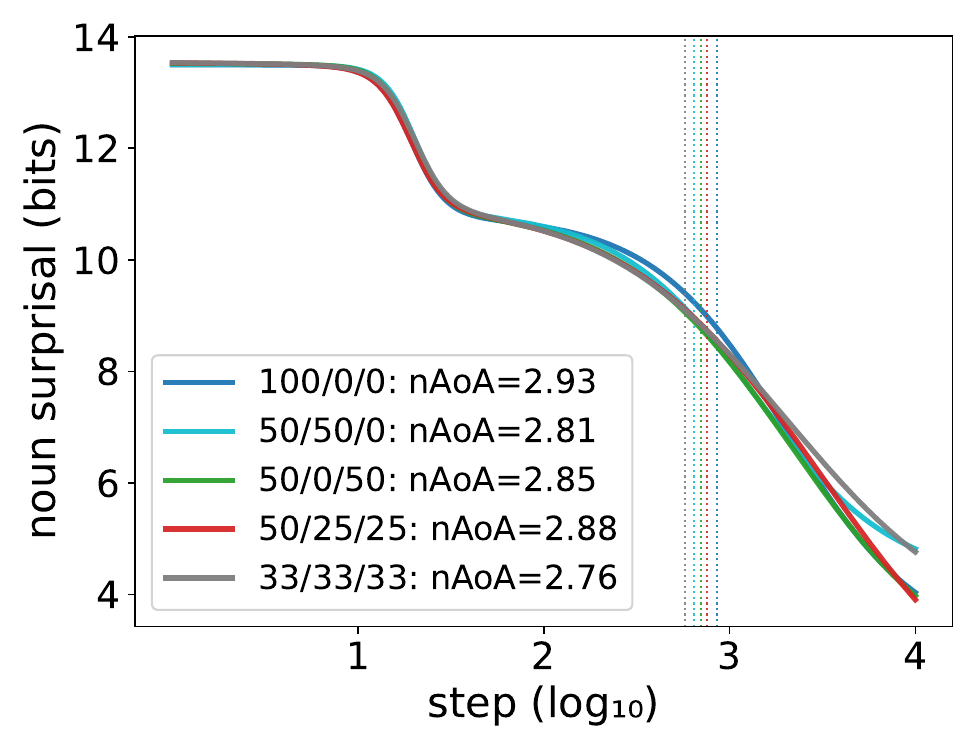}{1.0}{1.00}{0pt}{0 0 0 0}{\simactcellwL}}
        {\simactcelltune{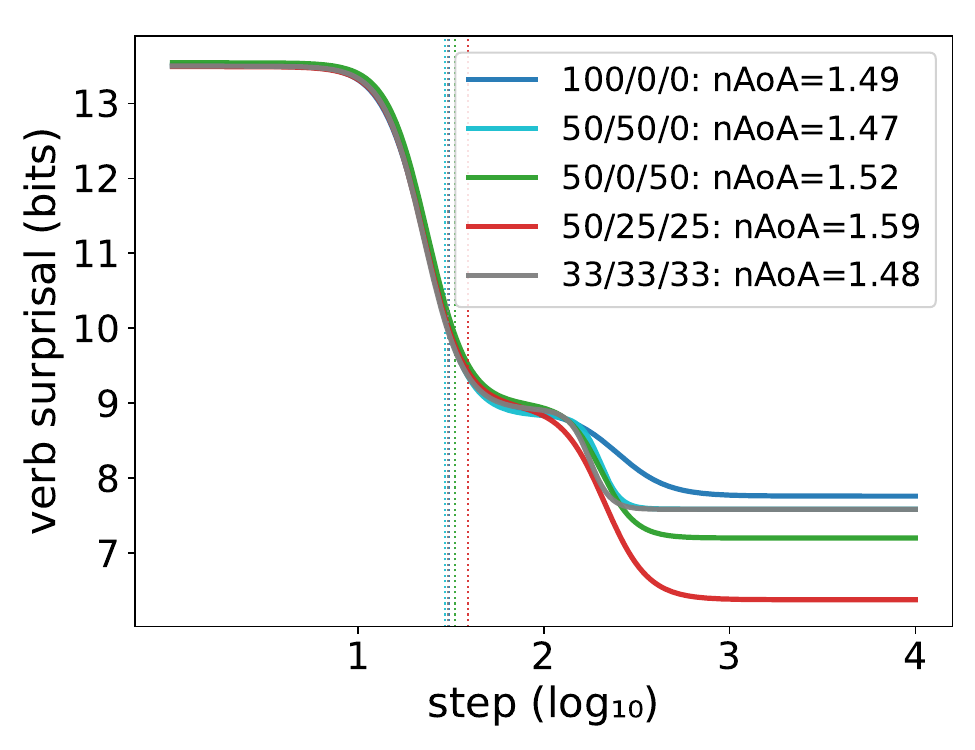}{1.0}{1.00}{0pt}{0 0 0 0}{\simactcellwL}}
        {\simactcelltune{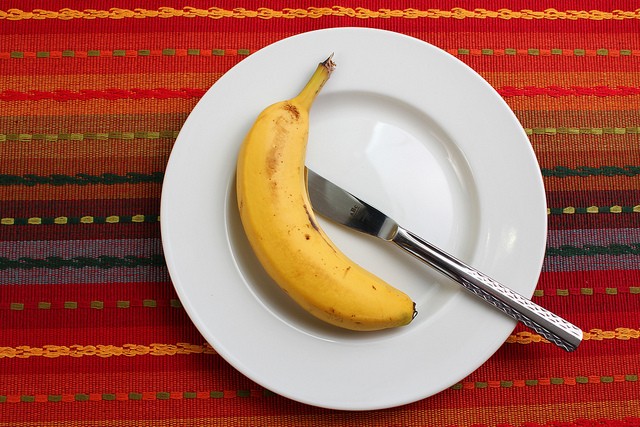}{0.95}{0.95}{5pt}{0 0 0 0}{\simactcellwR}}
        {\simactcelltune{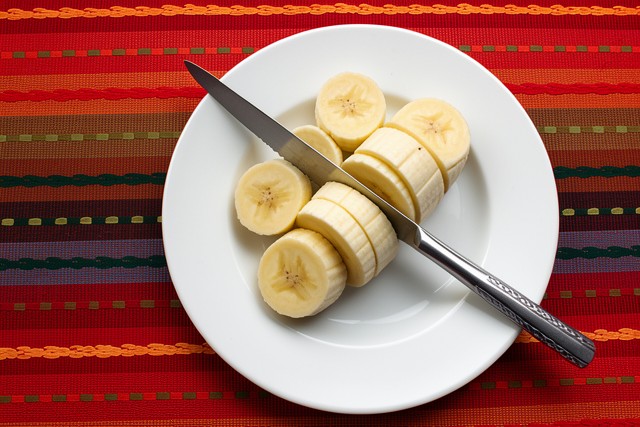}{0.95}{0.95}{5pt}{0 0 0 0}{\simactcellwR}}
        {Unseen pair: \textit{slice banana}.}
        {fig:simact-example-slice-banana}
    \simactrowcustom
        {\simactcelltune{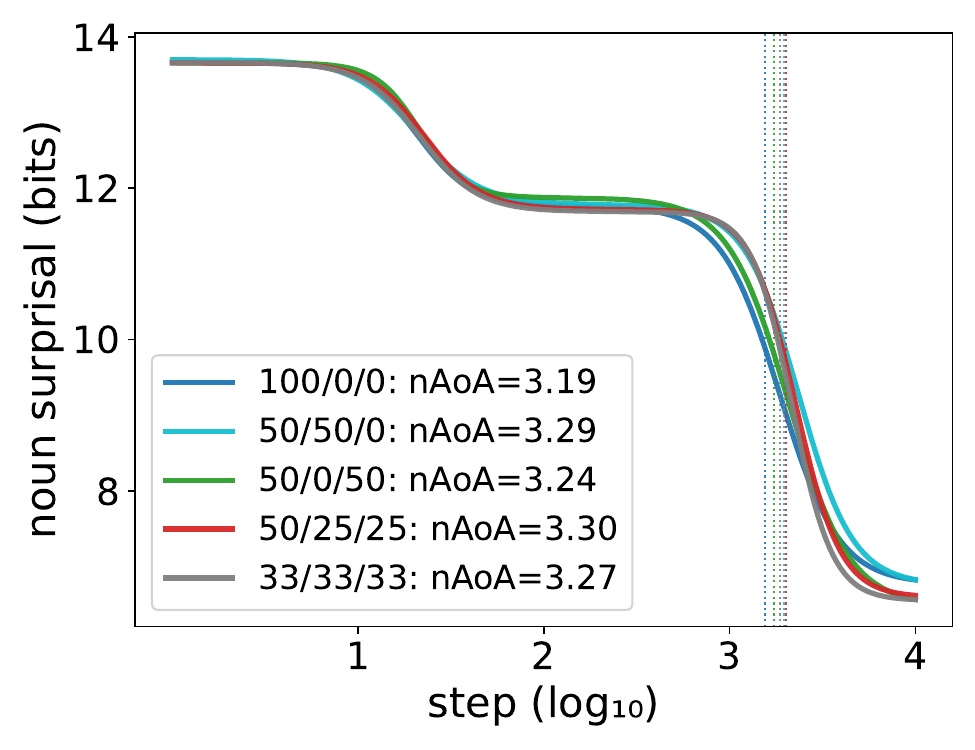}{1.00}{1.00}{0pt}{0 0 0 0}{\simactcellwL}}
        {\simactcelltune{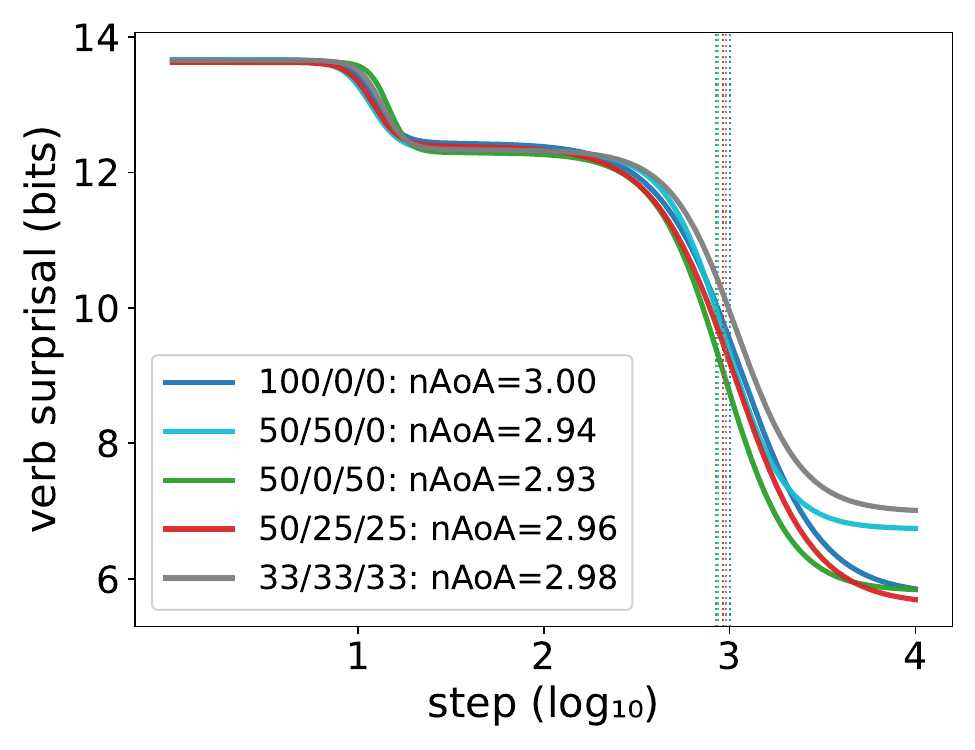}{1.00}{1.00}{0pt}{0 0 0 0}{\simactcellwL}}
        {\simactcelltune{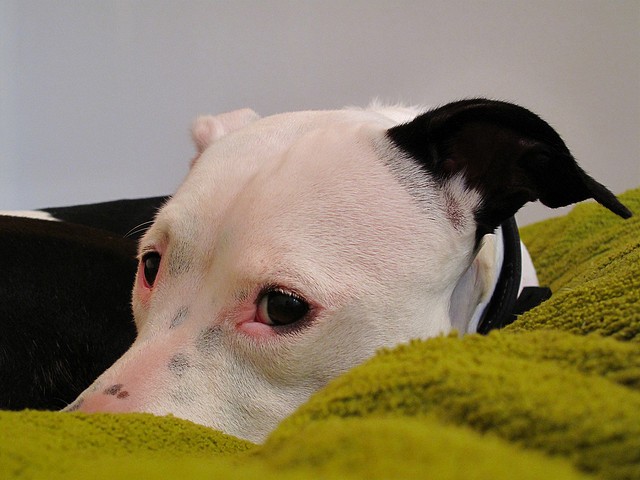}{0.92}{0.95}{5.5pt}{0 0 0 0}{\simactcellwR}}
        {\simactcelltune{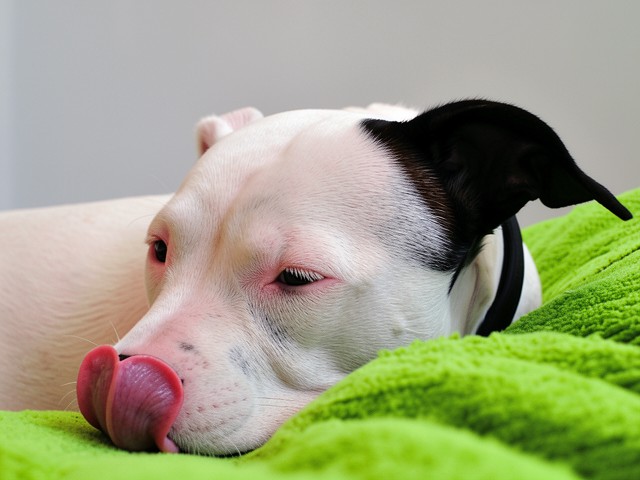}{0.92}{0.95}{5.5pt}{0 0 0 0}{\simactcellwR}}
        {Unseen pair: \textit{lick nose}.}
        {fig:simact-example-lick-nose}
    
    \vspace*{-5pt}
    \caption{Examples for two seen and two unseen verb-noun pairs. Rows correspond to seen result verb \textit{lower head}, seen manner verb \textit{pet dog}, unseen result verb \textit{slice banana}, and unseen manner verb \textit{lick nose}. Columns show the noun/verb surprisal learning curves and the pre-condition/post-condition images. Curves pool 3 random seeds under the full-sequence setting with InfoNCE loss.}
    \vspace*{-12pt}
    \label{fig:simact-example-all}
\end{figure*}

%% file: sec/07-ablation.tex
\section{Ablation and Analysis}
\label{sec:ablation}

To better understand the sources of improvement from visual simulation, we conduct additional controlled experiments and sensitivity analyses. Since pre- and post-condition images can be visually similar, the benefits of simulation may arise from additional visual supervision, rather than from learning the relationship between actions and their outcomes. 
To distinguish these effects, we train an action-shuffled control for each simulation mixture.
We present the action-shuffle experiment below and the frame-replacement analysis in Appendix~\ref{app:frame-replacement}.

\textbf{Training.}
We construct an action-shuffled variant of each supervision mixture by reassigning action texts in $\mathrm{Sim}_{\mathrm{fwd}}$ and $\mathrm{Sim}_{\mathrm{bwd}}$ sequences.
Specifically, we group training examples by their shared pre-condition image and randomly reassign action texts within each group through a derangement that remains fixed throughout training, ensuring that no example retains its original action assignment.
LM sequences retain their original action texts.
This procedure disrupts the correspondence between actions and visual state transitions while preserving the visual inputs, image-prediction targets, and distribution of action texts associated with each pre-condition.
For each supervision mixture, we train both variants under identical architectural and optimization settings for 10,000 steps with three random seeds.

\textbf{Evaluation.}
We evaluate pairs of test examples that share a pre-condition but differ in their actions and outcomes.
For a target verb or noun $w$ at position $k$ in the action sequence, we compute the difference between its surprisal $s_k$ (defined in Equation~\ref{eq:surprisal}) under the swapped and correct post-conditions, keeping the pre-condition and preceding action prefix unchanged:
\begin{equation}
D(w) = s_k^{\mathrm{swapped}} - s_k^{\mathrm{correct}}.
\end{equation}
Higher values of $D(w)$ indicate greater sensitivity to the specific outcome observed.

\textbf{Results.}
As shown in Table~\ref{tab:action_shuffle}, the original models have a larger $D$ than their shuffled counterparts for both verbs and nouns across all supervision mixtures. 
The improvements are comparable for manner and result verbs, and the original model is ahead in every same-seed comparison. 
Since the shuffled models receive identical visual supervision, these results suggest that the benefits of simulation arise not merely from additional visual supervision, but also from the correct correspondence between actions and their visual outcomes (i.e., action-conditioned learning).

\input{tables/action-shuffle}

%% file: tables/action-shuffle.tex
\begin{table*}[t!]
\centering
\caption{
    Test-time swap $D(w)$ (bits; higher means more outcome-specific word prediction) at training step 10,000, mean $\pm$ std over three seeds. Seen and unseen test sets are pooled. Manner/Result columns split examples by the class of their verb.
    \vspace*{-8pt}
}
    \resizebox{0.9\textwidth}{!}{
    \begingroup
    \setlength{\tabcolsep}{2.5pt}
    \renewcommand{\arraystretch}{0.9}
    \begin{tabular}{ccccccccccc}
    \toprule
    \multicolumn{4}{c}{Supervision}
    & \multirow{2}{*}{Model}
    & \multicolumn{3}{c}{Verb $D(w)$ $\uparrow$}
    & \multicolumn{3}{c}{Noun $D(w)$ $\uparrow$} \\
    \cmidrule(r){1-4} \cmidrule(r){6-8} \cmidrule(r){9-11}
    LM$_\text{bwd}$ & LM$_\text{fwd}$ & Sim$_\text{bwd}$ & Sim$_\text{fwd}$
    & & All & Manner & Result & All & Manner & Result \\
    \midrule
    -- & 50\% & 50\% & -- & \texttt{Original} & $\sota{2.81}_{\pm.06}$ & $\sota{2.85}_{\pm.05}$ & $\sota{2.72}_{\pm.03}$ & $\sota{2.18}_{\pm.12}$ & $\sota{1.87}_{\pm.10}$ & $\sota{2.19}_{\pm.15}$ \\
    -- & 50\% & 50\% & -- & \texttt{Shuffled} & $1.53_{\pm.07}$ & $1.59_{\pm.09}$ & $1.48_{\pm.11}$ & $1.01_{\pm.18}$ & $0.88_{\pm.17}$ & $1.02_{\pm.19}$ \\
    \midrule
    -- & 50\% & -- & 50\% & \texttt{Original} & $\sota{1.44}_{\pm.05}$ & $\sota{1.46}_{\pm.06}$ & $\sota{1.41}_{\pm.07}$ & $\sota{0.91}_{\pm.18}$ & $\sota{0.76}_{\pm.16}$ & $\sota{0.94}_{\pm.20}$ \\
    -- & 50\% & -- & 50\% & \texttt{Shuffled} & $1.18_{\pm.02}$ & $1.22_{\pm.02}$ & $1.12_{\pm.05}$ & $0.61_{\pm.02}$ & $0.48_{\pm.01}$ & $0.64_{\pm.02}$ \\
    \midrule
    -- & 50\% & 25\% & 25\% & \texttt{Original} & $\sota{1.94}_{\pm.12}$ & $\sota{1.99}_{\pm.09}$ & $\sota{1.89}_{\pm.14}$ & $\sota{1.36}_{\pm.19}$ & $\sota{1.16}_{\pm.17}$ & $\sota{1.37}_{\pm.20}$ \\
    -- & 50\% & 25\% & 25\% & \texttt{Shuffled} & $1.41_{\pm.06}$ & $1.44_{\pm.03}$ & $1.40_{\pm.12}$ & $1.01_{\pm.26}$ & $0.84_{\pm.25}$ & $1.04_{\pm.27}$ \\
    \bottomrule
    \end{tabular}
    \endgroup}
\vspace*{-10pt}

\label{tab:action_shuffle}
\end{table*}

%% file: sec/02-background.tex
\section{Related Work}

\subsection{Causal perception and world modeling}

A long tradition in cognitive science treats human reasoning as grounded in internal models of the world: structured representations that support prediction, counterfactual projection, causal explanation, and action planning~\citep{johnson1983mental,nersessian1993theoretician,wooldridge1995intelligent,kappes2016mental,russell2016artificial}. 
Developmental evidence suggests that this event structure emerges early: infants distinguish causal from non-causal events, represent event roles before they have a mature lexicon~\citep{leslie1987six,spelke1990principles,mandler1992build,oakes1990infant,cohen1998development}.
Since these roles are also central to verb meaning~\citep{cohen1999perceptual,solstad2017causality}, causal perception therefore links visual event understanding to action-word learning.


Computer vision has studied related questions through object affordances, human-object interactions, and causal relations between actions and object states~\citep{koppula2013learning,fathi2013modeling,yang2013detection,wang2016actions,zhuo2019explainable}. 
Other work infers object states from observations~\citep{fire2015learning,fire2017inferring,alayrac2017joint,liu2017joint,hong2021transform}, predicts future state changes~\citep{zhou2016learning,xiong2018learning,gao2018action}, or jointly models physical state and dynamics~\citep{wu2015galileo,wu2016physics,wu2017learning}. 
More recent world-modeling approaches further emphasize prediction as a central objective for agents.

Despite this progress, most prior work treats state change primarily as a visual or physical modeling problem, leaving the connection to lexical acquisition less explored. 

\subsection{Grounded word acquisition}

Grounded word learning has traditionally been framed as the problem of mapping words to referents in the world~\citep{harnad1990symbol,gleitman1994acquisition,clark1995lexicon}. 
Early computational and visual grounding work modeled word-to-symbol mappings, word-object alignment, and grounding in richer visual features and structured scene representations~\citep{siskind1996computational,regier2005emergence,goodman2007bayesian,fazly2010probabilistic,roy2002learning,yu2005emergence,xu2007word,yu2007unified,yu2013grounded,qu2010context,jiayuan2019neuro,mao2021grammarbased,pratt2020grounded}.
Large-scale vision-language models have extended this paradigm to region-level~\citep{li2022grounded,ma2023world,chen2023shikra,you2024ferret,wang2024cogvlm} and pixel-level~\citep{xia2023gsva,rasheed2024glamm,zhang2024groundhog} grounding, with strong performance on referring expression comprehension~\citep{chen2024revisiting}.

Despite this progress, much of this literature remains object-centered. 
This formulation is well suited to nouns but incomplete for verbs, which often denote event structure over time. 
Developmental work indicates that objects are stable perceptual units, whereas verbs require the learner to identify event components~\citep{gentner1982nouns,gentner2006verbs,golinkoff1987eyes}.
Verb learning is also shaped by the nouns that participate in the event~\citep{arunachalam2011grammatical,ferguson2014infants}. 
Together, these findings suggest that lexical acquisition couples object identity with action structure.

Our work studies this coupling in neural models.
Rather than treating grounding as static word-referent alignment, we ask whether lexical learning improves under action-conditioned visual transitions. 
This framing suggests that transition prediction is useful for verbs, especially result verbs. 

\subsection{Psychologically motivated analysis of language models}

Neural language models differ from human learners in data, objectives, architectures, and inductive biases~\citep{baroni2022proper,warstadt2022artificial}. 
Nevertheless, they can serve as controlled computational systems for testing which learning signals support particular linguistic behaviors. 
Prior work has connected language-model surprisal to psycholinguistic variables and lexical acquisition patterns~\citep{portelance2020predicting,ma2023world}, and has examined how linguistic knowledge emerges over training time~\citep{sellam2021multiberts,blevins2022analyzing,biderman2023pythia,xia2023training,chang2022word,chang2023characterizing,evanson2023language,ma2025babysit}.

If models are used as such scientific instruments, benchmark outcomes alone are insufficient~\citep{baroni2022proper,portelance2022neural}. 
The relevant question is which training factors make success possible~\citep{warstadt2022artificial,portelance2022neural,portelance2023roles}.
We therefore vary visual simulation supervision, measure lexical surprisal over training, and compare acquisition dynamics across nouns, verbs, manner verbs, and result verbs. 

%% file: sec/06-conclusion.tex
\section{Discussion and Conclusion}
This paper uses neural models as controlled learners to study whether learning to predict how actions transform visual states can facilitate the acquisition of action-word meanings.
Our results show that the benefits of predictive learning depend on lexical semantics: verbs benefit more than nouns, while result verbs exhibit greater improvements in acquisition speed than manner verbs. 

The implications of this work are two-fold. The first is from a language acquisition perspective. Cognitive studies have shown that in the first year of life, pre-linguistic infants can distinguish whether an event is causal or non-causal, predict event outcomes based on the ontological status of the objects involved, and recognize different action roles~\citep{leslie-infant,oakes1990infant,cohen1998development}. One significant tie between causal perception of actions and language lies between the perceived action roles (e.g., agent and patient) and the semantic roles of verb arguments~\citep{leslie-infant-perception,solstad2017causality}. These findings suggest a potential link between early word acquisition and causal perception, which motivates our work. Here we use neural models as controlled learners to study word acquisition under different training objectives of simulating the world. We do not claim our model accounts for human learning, but we hope that our initial findings will motivate future work to explore this link, e.g., by training and testing on more developmentally plausible data or conducting human studies. 

The second implication is from an AI perspective. We see two lines of work in building large AI models: (1) world models that learn from vision and simulation, with little language; (2) language models that learn language from linguistic/visual context, but without simulating the world. Although our current focus is on (2) where we show how predictive world modeling contributes to grounded language acquisition, we hope our initial results will motivate future work that will bring these two lines together to build foundations for embodied agents that can perceive, act, and communicate with humans.

\section{Limitations}

Our framework models single-step action-effect transitions and therefore does not yet test longer-horizon compositional reasoning over sequences of state changes. The controlled transition corpus is also generated using language and image generative models, so its action distribution inevitably reflects the generator's priors over plausible actions, participants, and resulting states. On the language side, we use a fixed-vocabulary whitespace tokenizer so that target verbs and nouns map to single tokens, which simplifies lexical surprisal evaluation but precludes subword generalization to related or unseen forms. Finally, our experiments use English captions and action descriptions only, and therefore do not address cross-linguistic variation in how events and state changes are lexicalized.

%% file: appendix/appendix.tex
\input{appendix/a1-method}

\input{appendix/a2-dataset}

\input{appendix/a3-experiment}

%% file: appendix/a1-method.tex
\section{Method}

\subsection{Architecture}
\label{app:architecture}

 We initialize the visual backbone from NEPA~\citep{xu2025next}, an autoregressive visual model pretrained with causal patch-embedding prediction, while initializing the language-specific components from scratch. 
This gives the model visual predictive structure at initialization, but requires it to acquire the mapping between words and visually grounded state changes during training.

\paragraph{Tokenization.}
We use whole-word rather than subword tokenization because the object of study is word learning, and the relevant lexical units are complete words rather than subword fragments.

\paragraph{Backbone.}
We use the \texttt{NEPA-large} backbone, which has 24 Transformer layers, hidden size 1024, and 16 attention heads, pretrained on ImageNet-1K. 
Image inputs are $224 \times 224$ tensors and are mapped to $N=256$ patch embeddings by a convolutional patch embedding layer. 
Text inputs are represented with whole-word tokenization over the corpus. 
The original NEPA model uses two-dimensional rotary positional embeddings tied to image-grid coordinates. 
Because our sequences interleave image patches, text tokens, and boundary tokens, we replace these with learned one-dimensional absolute positional embeddings shared across all sequence positions. 
We also add token-type embeddings to distinguish image patches, text tokens, and boundary tokens. 
The Transformer trunk is shared across modalities and sequence formats.

\paragraph{Output heads.}
The model has two output heads. 
The language head predicts the next whole-word token and is tied to the text embedding matrix~\citep{radford2019language}. 
The image head is a linear projection that predicts the next visual patch embedding. 
We initialize the patch embedding layer, Transformer trunk, and final layer normalization from \texttt{NEPA-large}.
The text embeddings, token-type embeddings, absolute positional embeddings, and language head are randomly initialized.

\subsection{Sequence Formats}
\label{app:sequence}

Each training example contains a pre-condition image, an action text, and a post-condition image. 
At load time, we randomly sample one sequence format for each example according to fixed probabilities. 
All formats share the same model parameters and the same causal next-element prediction objective. 
They differ only in the order of the visual and textual components and in which positions are included in the loss. 
Special tokens \texttt{<pre>}, \texttt{</pre>}, \texttt{<post>}, and \texttt{</post>} mark the beginning and end of the pre-condition and post-condition image blocks in the actual input sequence.

\paragraph{Forward simulation.}
In the forward simulation format, the pre-condition image and action text appear before the post-condition image. 
This corresponds to \textbf{pre-condition $+$ action $\rightarrow$ post-condition}: the model predicts the resulting visual state from an initial state and an action description.

\paragraph{Backward simulation.}
In the backward simulation format, the post-condition image and action text appear before the pre-condition image. 
This corresponds to \textbf{post-condition $+$ action $\rightarrow$ pre-condition}: the model infers a compatible prior visual state from an observed result and an action description. 
Like forward simulation, this objective grounds action words in visually represented state changes; empirically, it can provide a stronger learning signal than forward simulation.

\paragraph{Forward language modeling.}
In the forward language modeling format, the pre-condition and post-condition images appear before the action text. 
This corresponds to \textbf{pre-condition $+$ post-condition $\rightarrow$ action}: the model predicts the action text associated with an observed visual transition, i.e., video captioning.

\paragraph{Backward language modeling.}
In the backward language modeling format, the post-condition and pre-condition images appear before the action text. 
This corresponds to \textbf{post-condition $+$ pre-condition $\rightarrow$ action}. 
We use this to reject the alternative hypothesis that improvements come from the ordering of visual states in language modeling, rather than from visual simulation objectives that require image-embedding prediction.

\paragraph{Format sampling.}
The sampling probabilities over sequence formats determine how the training budget is allocated between language modeling and visual simulation. 
A pure language-modeling baseline assigns all examples to language-modeling formats, while simulation-augmented settings reserve part of the budget for image-embedding prediction in one or both directions. 
Our default setting assigns 50\% of examples to language modeling and 50\% to simulation objectives. 
In Section~\ref{sec:findings}, we vary this allocation to measure how different mixtures of forward language modeling, backward language modeling, forward simulation, and backward simulation affect word learning.

%% file: appendix/a2-dataset.tex
\section{Dataset}
\label{app:dataset}

\subsection{World-model-distilled dataset}

\paragraph{Generation procedure.}
Our generation pipeline begins with an MSCOCO 2017 source image and requests four to five candidate future actions from GPT-5.4-nano. 
For each candidate, the structured response contains an action phrase, its moving subjects, an expected visual change, an action type, an optional viewpoint change, and a corresponding image-editing instruction.
The five action types are \emph{object interaction}, \emph{body motion}, \emph{scene transition}, \emph{state change}, and \emph{new element}. The prompt requests at least three action types, at least one new-element candidate, and at least two event-linked viewpoint changes per source. It emphasizes physically and temporally plausible, visibly distinguishable outcomes and preservation of relevant object identities, materials, support relations, and scene content. These are prompting constraints, rather than independently measured properties of every generated sample.

Each candidate's editing instruction is applied independently to the same source image with FLUX.2 [klein]-9B.
The action phrase becomes the training text, while the expected-change field is retained as auxiliary metadata. This field is predicted before image editing and is not a separate caption of the realized output image.

\paragraph{Image-generation settings.}
Table~\ref{tab:appendix_flux_settings} reports the generation configuration.
We preserve the source canvas by padding the right and bottom edges to multiples of 16 using edge pixels, generate the edited image, and crop back to the original dimensions. The implementation accepts inputs with both dimensions at least 64 pixels and a total area at most four megapixels.

\begin{table}[t]
\centering
\caption{Parameters in the available generation code and run templates.}
\small
\begin{tabular}{ll}
\toprule
Parameter & Available configuration \\
\midrule
Model & FLUX.2 [klein]-9B \\
Denoising steps & 4 \\
Guidance scale & 1.0 \\
Edited images per action & 1 \\
\bottomrule
\end{tabular}

\label{tab:appendix_flux_settings}
\end{table}

\paragraph{Structured validation and curation.}
The pipeline validates the generated JSON structure and requires a nonempty action phrase, expected-change description, and exactly one editing prompt for each action index. Invalid batch responses are logged and rerun. 

\paragraph{Splits and lexical metadata.}
Table~\ref{tab:appendix_synthetic_counts} describes the statistics of the world-model-distilled dataset. The two test subsets used for the main experiments are curated from the full test pool.

\paragraph{Vocabulary counting.}
Vocabulary sizes count distinct non-missing annotation labels within each split. Verb counts use the verb labels.
Noun vocabularies combine the subject and object labels.
Annotation strings are counted without additional normalization. Vocabulary sizes pooled across sources are computed from the union of their label sets.

\paragraph{Illustrative generated pairs.}
Figure~\ref{fig:appendix_generated_examples} shows eight generated pairs sampled from the world-model-distilled dataset.

\input{tables/appendix/table_synthetic}
\input{figures/appendix/data_qualitative}

\subsection{Real-video dataset}

\paragraph{Sources and temporal boundaries.}
The corpus pools Assembly101, Ego4D FHO-LTA, EPIC-KITCHENS-100, FineAction, and HoloAssist. An example uses the annotated segment's start and end boundaries as the pre-condition and post-condition observations. The extraction retains each source's timestamp or frame-index convention.

\paragraph{Action annotation and splits.}
Each action is described by an image-conditioned Gemini sentence and paired with cleaned verb and object labels. Lexical alignment provides token positions and matched surface forms.
Table~\ref{tab:appendix_real_counts} gives the statistics for each source and
split.

\input{tables/appendix/table_real_video}

\paragraph{Manner and result inventory.}
The operational classification uses a manually curated lemma inventory.
Manner verbs describe how an action is performed, such as its characteristic motion or instrument. The diagnostic permits the result to be left implicit. Result verbs specify a resulting change of state, location, or possession. 
Classification is a lemma-level lookup rather than a contextual semantic classifier. Consequently, the labels should be interpreted as the study's operational choices, especially for polysemous verbs.

The complete inventory contains 73 manner, 76 result, and 23 ambiguous entries.

\input{tables/appendix/selected_verbs}

\paragraph{Vocabulary counting.}
Vocabulary sizes count distinct non-missing annotation labels within each split. Verb counts use the verb labels.
Noun vocabularies use the object labels.
Annotation strings are counted without additional normalization. Vocabulary sizes pooled across sources are computed from the union of their label sets.

%% file: tables/appendix/table_synthetic.tex
\begin{table}[t]
\centering
\caption{World-model-distilled statistics. Seen and unseen are curated subsets of the full test pool. Nouns count distinct stored subject and object labels jointly, excluding missing values.}
\small
\begin{tabular}{lrrr}
\toprule
Split & Triples & Verbs & Nouns \\
\midrule
Train & 488,703 & 1,788 & 5,322 \\
Validation & 20,696 & 785 & 1,924 \\
Full test pool & 168,690 & 1,338 & 3,863 \\
\midrule
Test (seen) & 47,332 & 97 & 1,220 \\
Test (unseen) & 12,097 & 120 & 794 \\
\bottomrule
\end{tabular}
\label{tab:appendix_synthetic_counts}
\end{table}

%% file: figures/appendix/data_qualitative.tex
\begin{figure}[t]
\centering
\includegraphics[width=\textwidth,height=.86\textheight,keepaspectratio]{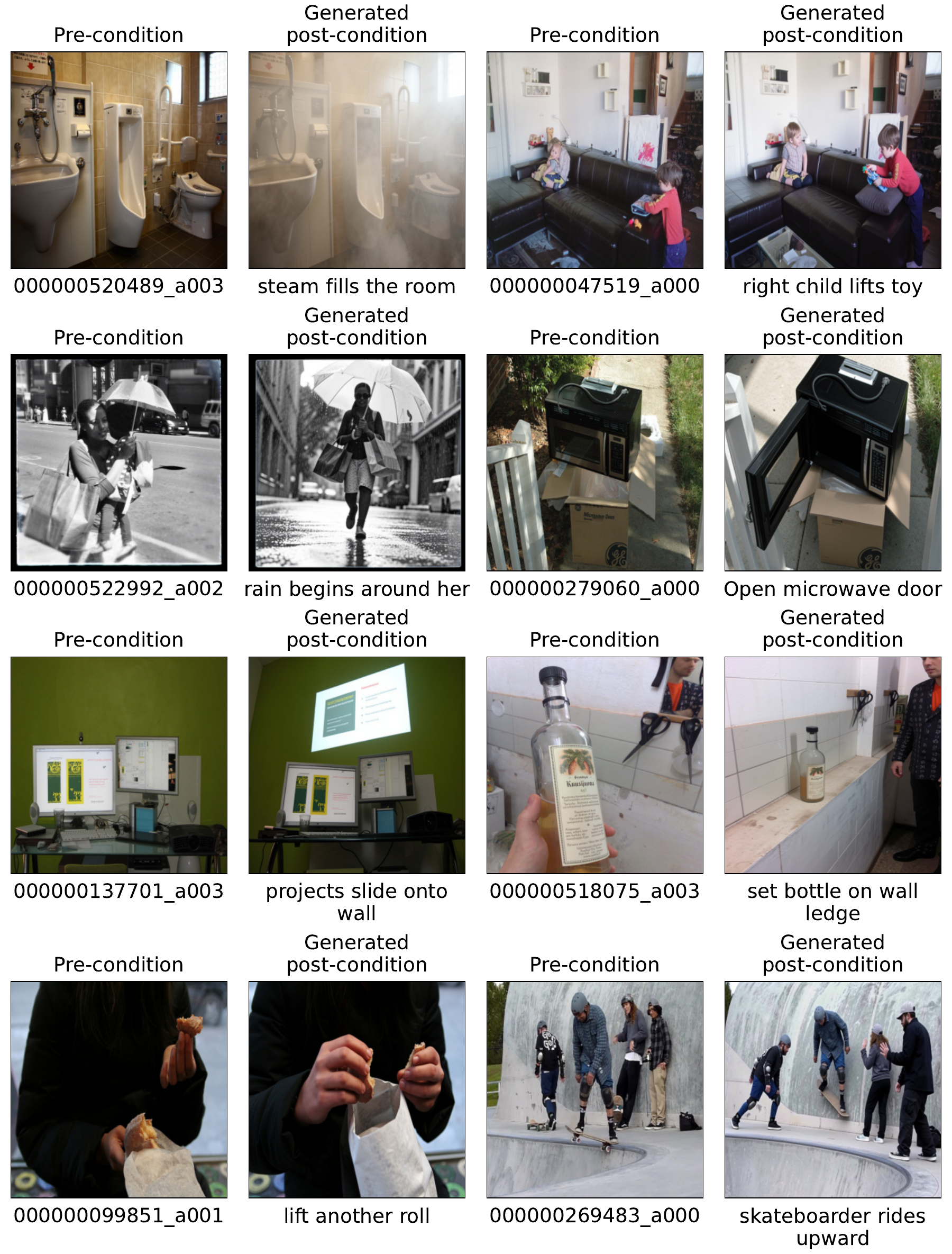}
\caption{Sampled pre-/post-condition pairs in the world-model-distilled dataset. Each pair shows the source on the left and the generated state on the right.}
\label{fig:appendix_generated_examples}
\end{figure}

%% file: tables/appendix/table_real_video.tex
\begin{table}[t]
\centering
\caption{Real-video statistics by source and split. Train and Test (curated) use the same split definitions as the main-text table. Test (curated) is a 3,999-example subset of the full evaluation pool. Vocabulary sizes count distinct nonmissing verb and object labels within each row; All sources takes the union across sources within that split. The full evaluation pool includes 95,690 source-validation examples and 21,893 Assembly101 official-test examples.}
\small
\begin{tabular}{llrrr}
\toprule
Source & Split & Triples & Verbs & Nouns \\
\midrule
Assembly101 & Train & 47,535 & 16 & 58 \\
 & Test (curated) & 393 & 11 & 38 \\
 & Full eval. pool & 37,565 & 16 & 60 \\
\midrule
Ego4D FHO-LTA & Train & 63,846 & 116 & 460 \\
 & Test (curated) & 1,598 & 46 & 247 \\
 & Full eval. pool & 33,001 & 108 & 419 \\
\midrule
EPIC-KITCHENS-100 & Train & 67,109 & 342 & 699 \\
 & Test (curated) & 457 & 32 & 129 \\
 & Full eval. pool & 9,646 & 168 & 359 \\
\midrule
FineAction & Train & 47,852 & 45 & 64 \\
 & Test (curated) & 990 & 32 & 49 \\
 & Full eval. pool & 19,474 & 44 & 63 \\
\midrule
HoloAssist & Train & 124,702 & 45 & 102 \\
 & Test (curated) & 561 & 22 & 59 \\
 & Full eval. pool & 17,897 & 46 & 98 \\
\midrule
All sources & Train & 351,044 & 400 & 1,016 \\
 & Test (curated) & 3,999 & 80 & 407 \\
 & Full eval. pool & 117,583 & 257 & 724 \\
\bottomrule
\end{tabular}
\label{tab:appendix_real_counts}
\end{table}

%% file: tables/appendix/selected_verbs.tex
\begingroup

\noindent\textbf{Manner (73).} bat, blow, braid, brush, carry, carve, cast, chop, clap, click, dip, draw, dribble, drink, drizzle, dunk, flick, fly, hit, hold, inspect, kick, knead, lick, mop, mow, offer, pass, pet, pitch, play, pour, press, puff, pull, push, reel, ride, rinse, rotate, rub, run, sand, scoop, scrape, scratch, serve, sew, shake, shave, shovel, signal, skip, slap, slide, smear, spin, spray, squeeze, step, stir, stretch, strike, sweep, swing, tap, throw, toss, touch, walk, wash, weld, wipe.\par

\noindent\textbf{Result (76).} add, adjust, align, appear, approach, arrange, assemble, attach, board, break, clean, close, come, complete, connect, cover, cross, cut, detach, disassemble, divide, drape, drop, dry, empty, enter, fill, flush, fog, fold, get, hang, insert, lead, light, lower, mark, mold, mount, open, pack, paste, peel, place, plant, plug, position, power, put, raise, reach, regain, release, remove, repair, retract, retrieve, secure, set, shift, shut, sit, slice, smooth, straighten, switch, take, tighten, tilt, transfer, trim, tuck, unfold, unscrew, water, withdraw.\par

\noindent\textbf{Ambiguous (23).} apply, attempt, catch, climb, dig, flip, grab, keep, lift, make, measure, mix, move, off, on, operate, paint, pick, roll, screw, shoot, tie, turn.\par

\endgroup

%% file: appendix/a3-experiment.tex
\section{Experiment}
\label{app:experiment}

\subsection{Baselines and Ablation Variants}
\label{sec:baseline-ablation}

To reliably assess the contribution of each design choice in our framework, we conduct controlled ablation studies.
We describe our baselines and variants below. 

\vspace*{-5pt}
\begin{itemize}[leftmargin=*]
    \setlength\itemsep{-0.25em}

    \item \(\mathrm{LM}_{\mathrm{fwd}} + \mathrm{Sim}_{\mathrm{bwd}} + \mathrm{Sim}_{\mathrm{fwd}}\): the main simulation-augmented setting, with 50\% forward language modeling, 25\% backward simulation, and 25\% forward simulation. 
    This tests word learning when action-text prediction is jointly trained with visual simulation in both directions.

    \item \(\mathrm{LM}_{\mathrm{fwd}} + \mathrm{Sim}_{\mathrm{bwd}}\): removes forward simulation and keeps backward simulation. 
    This isolates the contribution of inverse visual prediction.

    \item \(\mathrm{LM}_{\mathrm{fwd}} + \mathrm{Sim}_{\mathrm{fwd}}\): removes backward simulation and keeps forward simulation. 
    This isolates the contribution of future-state visual prediction.

    \item \(\mathrm{LM}_{\mathrm{fwd}}\) only: removes both simulation objectives. 
    This is the main LM-only baseline, testing word learning without action-conditioned visual state prediction.

    \item \(\mathrm{LM}_{\mathrm{bwd}}\) only: uses backward language modeling without action-conditioned visual state prediction.
    This controls for whether reversing the order of the visual states changes the language-only baseline.

    \item \(\mathrm{LM}_{\mathrm{fwd}} + \mathrm{LM}_{\mathrm{bwd}}\): uses both language-modeling directions without simulation. 
    This tests whether exposure to both visual-state orders helps independently of action-conditioned visual state prediction.
\end{itemize}
\vspace*{-5pt}
For each variant, we train on each corpus with 3 random seeds for 10{,}000 steps.
Unless otherwise noted, we treat full-sequence supervision with the InfoNCE image loss as our main setting and report contrasts with the cosine and completion-only variants only when they alter the qualitative picture.

\subsection{Preprocessing}
\label{app:preprocessing}

Both corpora are converted to the same training format.
All images are resized to $224 \times 224$ and normalized to $[-1,1]$ before patch embedding.
Each triple is rendered into one of the sequence formats defined in Section~\ref{sec:sequence-format}; the sampling probabilities over formats are determined by the training variant in Section~\ref{sec:baseline-ablation}.
Models differ only in the objective mixture, image loss, or supervision mask.

All variants use NEPA-large, with 24 transformer layers, a hidden dimension of 1,024, and 16 attention heads. The patch embedding, transformer, and final layer normalization are initialized from the same pretrained NEPA checkpoint.
Word embeddings and one-dimensional positional embeddings are newly initialized, and the language-model output weights are tied to the word embeddings. 
Within each corpus, comparisons share the training split, vocabulary, preprocessing, and model architecture. Variants differ in sequence-format sampling, image loss, or supervision mask.

\subsection{Frame-Replacement Control}
\label{app:frame-replacement}

\input{tables/appendix/frame-replacement}

To investigate why backward simulation yields greater improvements than forward simulation on the world-model-distilled dataset, we examine how models trained with different simulation objectives rely on the pre- and post-condition images.

\textbf{Experiment setup.}
We evaluate the models at the training step 10{,}000 using the $\mathrm{LM}_{\mathrm{fwd}}$ format $[I_{\mathrm{pre}}, I_{\mathrm{post}}, a]$. 
We replace either $I_{\mathrm{pre}}$ or $I_{\mathrm{post}}$ with the corresponding image from a randomly sampled test example, while keeping the other image and the action text $a$ unchanged. 
We measure the increase in verb and noun surprisal, with higher values indicating greater reliance on the replaced image.

\textbf{Backward simulation models rely more heavily on post-conditions.}
Table~\ref{tab:frame_replacement} shows that models trained with backward simulation are substantially more sensitive to the replacement of post-condition images than pre-condition images.
In contrast, models trained with forward simulation rely on both images approximately equally for verb prediction and more heavily on the pre-condition image for noun prediction.
Models trained with bidirectional simulation exhibit an intermediate pattern.

\textbf{The backward advantage is smaller on real videos.}
The advantage of backward over forward simulation is more pronounced on the world-model-distilled dataset (Table~\ref{tab:results}) than on real videos (Table~\ref{tab:video-results}). 
Our generated data tends to present more canonical action outcomes, as the post-condition images are explicitly constructed to depict recognizable consequences of the described actions. 
In contrast, the effects of actions in real videos are often less clearly isolated.
Accordingly, backward and forward simulation achieve comparable performance on real videos, while both improve over LM-only training. These findings suggest that more canonical action outcomes can amplify the benefits of backward simulation, while the improvements on real videos demonstrate that our approach remains effective under less idealized conditions.

%% file: tables/appendix/frame-replacement.tex
\begin{table*}[!t]
\centering
\caption{
    Frame replacement in the $\mathrm{LM}_{\mathrm{fwd}}$ format $[I_{\mathrm{pre}}, I_{\mathrm{post}}, a]$ on the 10k checkpoints, mean over three seeds. Each cell is the increase in word surprisal (bits) when the pre-condition (Pre) or post-condition (Post) image is replaced with the same-type image of a random other test example; higher means greater reliance on the replaced image. Bold values indicate greater reliance within each Pre/Post comparison.
    \vspace*{-8pt}
}
    \resizebox{0.8\textwidth}{!}{
    \begingroup
    \setlength{\tabcolsep}{2.5pt}
    \renewcommand{\arraystretch}{0.9}
    \begin{tabular}{cccccccccccc}
    \toprule
    \multicolumn{4}{c}{Supervision}
    & \multicolumn{2}{c}{Verb, Seen}
    & \multicolumn{2}{c}{Verb, Unseen}
    & \multicolumn{2}{c}{Noun, Seen}
    & \multicolumn{2}{c}{Noun, Unseen} \\
    \cmidrule(r){1-4} \cmidrule(r){5-6} \cmidrule(r){7-8} \cmidrule(r){9-10} \cmidrule(r){11-12}
    $\mathrm{LM}_{\mathrm{bwd}}$ & $\mathrm{LM}_{\mathrm{fwd}}$ & $\mathrm{Sim}_{\mathrm{bwd}}$ & $\mathrm{Sim}_{\mathrm{fwd}}$
    & Pre & Post & Pre & Post & Pre & Post & Pre & Post \\
    \midrule
    -- & 50\% & 50\% & -- & $0.16$ & $\sota{5.04}$ & $0.20$ & $\sota{3.85}$ & $0.22$ & $\sota{5.87}$ & $0.25$ & $\sota{5.71}$ \\
    -- & 50\% & -- & 50\% & $\sota{1.85}$ & $1.76$ & $\sota{1.62}$ & $\sota{1.62}$ & $\sota{2.37}$ & $1.22$ & $\sota{2.94}$ & $1.13$ \\
    -- & 50\% & 25\% & 25\% & $0.61$ & $\sota{2.93}$ & $0.68$ & $\sota{2.58}$ & $0.86$ & $\sota{2.64}$ & $1.13$ & $\sota{2.69}$ \\
    \bottomrule
    \end{tabular}
    \endgroup}
\vspace*{-9pt}

\label{tab:frame_replacement}
\end{table*}

%% file: references.bib
@misc{gpt4o,
    url       = {https://openai.com/index/hello-gpt-4o/},
    title     = {Hello GPT-4o},
    author    = {OpenAI},
    publisher = {OpenAI},
    year      = {2024},
    month     = {May}
}

@misc{anthropic2024claude,
    url       = {https://www.anthropic.com/news/claude-3-family},
    title     = {The Claude 3 Model Family: Opus, Sonnet, Haiku},
    author    = {Anthropic},
    publisher = {Anthropic},
    year      = {2024},
    month     = {March}
}

@inproceedings{li2022grounded,
    title     = {Grounded language-image pre-training},
    author    = {Li, Liunian Harold and Zhang, Pengchuan and Zhang, Haotian and Yang, Jianwei and Li, Chunyuan and Zhong, Yiwu and Wang, Lijuan and Yuan, Lu and Zhang, Lei and Hwang, Jenq-Neng and others},
    booktitle = {Proceedings of the IEEE/CVF Conference on Computer Vision and Pattern Recognition},
    pages     = {10965--10975},
    year      = {2022}
}

@inproceedings{ma2023world,
    title     = {World-to-Words: Grounded Open Vocabulary Acquisition through Fast Mapping in Vision-Language Models},
    author    = {Ma, Ziqiao and Pan, Jiayi and Chai, Joyce},
    booktitle = {Proceedings of the 61st Annual Meeting of the Association for Computational Linguistics (Volume 1: Long Papers)},
    pages     = {524--544},
    year      = {2023}
}

@article{wang2024cogvlm,
    title   = {Cogvlm: Visual expert for pretrained language models},
    author  = {Wang, Weihan and Lv, Qingsong and Yu, Wenmeng and Hong, Wenyi and Qi, Ji and Wang, Yan and Ji, Junhui and Yang, Zhuoyi and Zhao, Lei and XiXuan, Song and others},
    journal = {Advances in Neural Information Processing Systems},
    volume  = {37},
    pages   = {121475--121499},
    year    = {2024}
}

@inproceedings{you2024ferret,
    title     = {Ferret: Refer and Ground Anything Anywhere at Any Granularity},
    author    = {You, Haoxuan and Zhang, Haotian and Gan, Zhe and Du, Xianzhi and Zhang, Bowen and Wang, Zirui and Cao, Liangliang and Chang, Shih-Fu and Yang, Yinfei},
    booktitle = {The Twelfth International Conference on Learning Representations},
    year      = {2024}
}

@inproceedings{zhang2024groundhog,
    title     = {GROUNDHOG: Grounding Large Language Models to Holistic Segmentation},
    author    = {Zhang, Yichi and Ma, Ziqiao and Gao, Xiaofeng and Shakiah, Suhaila and Gao, Qiaozi and Chai, Joyce},
    booktitle = {Proceedings of the IEEE/CVF Conference on Computer Vision and Pattern Recognition},
    year      = {2024}
}

@inproceedings{rasheed2024glamm,
    title     = {Glamm: Pixel grounding large multimodal model},
    author    = {Rasheed, Hanoona and Maaz, Muhammad and Shaji, Sahal and Shaker, Abdelrahman and Khan, Salman and Cholakkal, Hisham and Anwer, Rao M and Xing, Erix and Yang, Ming-Hsuan and Khan, Fahad S},
    booktitle = {Proceedings of the IEEE/CVF Conference on Computer Vision and Pattern Recognition},
    year      = {2024}
}

@article{siskind1996computational,
    title     = {A computational study of cross-situational techniques for learning word-to-meaning mappings},
    author    = {Siskind, Jeffrey Mark},
    journal   = {Cognition},
    volume    = {61},
    number    = {1-2},
    pages     = {39--91},
    year      = {1996},
    publisher = {Elsevier}
}

@article{harnad1990symbol,
    title     = {The symbol grounding problem},
    author    = {Harnad, Stevan},
    journal   = {Physica D: Nonlinear Phenomena},
    volume    = {42},
    number    = {1-3},
    pages     = {335--346},
    year      = {1990},
    publisher = {Elsevier}
}

@inproceedings{lin2014microsoft,
    title        = {Microsoft coco: Common objects in context},
    author       = {Lin, Tsung-Yi and Maire, Michael and Belongie, Serge and Hays, James and Perona, Pietro and Ramanan, Deva and Doll{\'a}r, Piotr and Zitnick, C Lawrence},
    booktitle    = {European conference on computer vision},
    pages        = {740--755},
    year         = {2014},
    organization = {Springer}
}

@book{gleitman1994acquisition,
    author    = {Gleitman, Lila R and Landau, Barbara},
    publisher = {MIT Press},
    title     = {The acquisition of the lexicon},
    year      = {1994}
}

@book{clark1995lexicon,
    author    = {Clark, Eve V},
    number    = {65},
    publisher = {Cambridge University Press},
    title     = {The lexicon in acquisition},
    year      = {1995}
}

@article{regier2005emergence,
    author    = {Regier, Terry},
    journal   = {Cognitive science},
    number    = {6},
    pages     = {819--865},
    publisher = {Wiley Online Library},
    title     = {The emergence of words: Attentional learning in form and meaning},
    volume    = {29},
    year      = {2005}
}

@article{goodman2007bayesian,
    author  = {Goodman, Noah and Tenenbaum, Joshua and Black, Michael},
    journal = {Advances in neural information processing systems},
    title   = {A Bayesian framework for cross-situational word-learning},
    volume  = {20},
    year    = {2007}
}

@article{fazly2010probabilistic,
    author    = {Fazly, Afsaneh and Alishahi, Afra and Stevenson, Suzanne},
    journal   = {Cognitive Science},
    number    = {6},
    pages     = {1017--1063},
    publisher = {Wiley Online Library},
    title     = {A probabilistic computational model of cross-situational word learning},
    volume    = {34},
    year      = {2010}
}

@article{roy2002learning,
    author    = {Roy, Deb K and Pentland, Alex P},
    journal   = {Cognitive science},
    number    = {1},
    pages     = {113--146},
    publisher = {Wiley Online Library},
    title     = {Learning words from sights and sounds: A computational model},
    volume    = {26},
    year      = {2002}
}

@article{yu2005emergence,
    author    = {Yu, Chen},
    journal   = {Connection science},
    number    = {3-4},
    pages     = {381--397},
    publisher = {Taylor \& Francis},
    title     = {The emergence of links between lexical acquisition and object categorization: A computational study},
    volume    = {17},
    year      = {2005}
}

@article{xu2007word,
    author    = {Xu, Fei and Tenenbaum, Joshua B},
    journal   = {Psychological review},
    number    = {2},
    pages     = {245},
    publisher = {American Psychological Association},
    title     = {Word learning as Bayesian inference.},
    volume    = {114},
    year      = {2007}
}

@article{yu2007unified,
    author    = {Yu, Chen and Ballard, Dana H},
    journal   = {Neurocomputing},
    number    = {13-15},
    pages     = {2149--2165},
    publisher = {Elsevier},
    title     = {A unified model of early word learning: Integrating statistical and social cues},
    volume    = {70},
    year      = {2007}
}

@inproceedings{yu2013grounded,
    author    = {Yu, Haonan and Siskind, Jeffrey Mark},
    booktitle = {Proceedings of the 51st Annual Meeting of the Association for Computational Linguistics (Volume 1: Long Papers)},
    pages     = {53--63},
    title     = {Grounded language learning from video described with sentences},
    year      = {2013}
}

@article{qu2010context,
    author  = {Qu, Shaolin and Chai, Joyce Yue},
    journal = {Journal of Artificial Intelligence Research},
    pages   = {247--277},
    title   = {Context-based word acquisition for situated dialogue in a virtual world},
    volume  = {37},
    year    = {2010}
}

@article{jiayuan2019neuro,
    author  = {Jiayuan Mao and Chuang Gan and Pushmeet Kohli and Joshua B. Tenenbaum and Jiajun Wu},
    journal = {International Conference on Learning Representations (ICLR)},
    title   = {The Neuro-Symbolic Concept Learner: Interpreting Scenes, Words, Sentences From Natural Supervision},
    year    = {2019}
}

@inproceedings{mao2021grammarbased,
    author    = {Jiayuan Mao and Freda H. Shi and Jiajun Wu and Roger P. Levy and Joshua B. Tenenbaum},
    booktitle = {Advances in Neural Information Processing Systems},
    editor    = {A. Beygelzimer and Y. Dauphin and P. Liang and J. Wortman Vaughan},
    title     = {Grammar-Based Grounded Lexicon Learning},
    year      = {2021}
}

@inproceedings{pratt2020grounded,
    author       = {Pratt, Sarah and Yatskar, Mark and Weihs, Luca and Farhadi, Ali and Kembhavi, Aniruddha},
    booktitle    = {European Conference on Computer Vision},
    organization = {Springer},
    pages        = {314--332},
    title        = {Grounded situation recognition},
    year         = {2020}
}

@article{chen2023shikra,
    title   = {Shikra: Unleashing Multimodal LLM's Referential Dialogue Magic},
    author  = {Chen, Keqin and Zhang, Zhao and Zeng, Weili and Zhang, Richong and Zhu, Feng and Zhao, Rui},
    journal = {arXiv preprint arXiv:2306.15195},
    year    = {2023}
}

@inproceedings{xia2023gsva,
    title     = {GSVA: Generalized Segmentation via Multimodal Large Language Models},
    author    = {Xia, Zhuofan and Han, Dongchen and Han, Yizeng and Pan, Xuran and Song, Shiji and Huang, Gao},
    booktitle = {Proceedings of the IEEE/CVF Conference on Computer Vision and Pattern Recognition},
    year      = {2024}
}

@article{chen2024revisiting,
    title   = {Revisiting referring expression comprehension evaluation in the era of large multimodal models},
    author  = {Chen, Jierun and Wei, Fangyun and Zhao, Jinjing and Song, Sizhe and Wu, Bohuai and Peng, Zhuoxuan and Chan, S-H Gary and Zhang, Hongyang},
    journal = {arXiv preprint arXiv:2406.16866},
    year    = {2024}
}

@inproceedings{sellam2021multiberts,
    title     = {The MultiBERTs: BERT Reproductions for Robustness Analysis},
    author    = {Sellam, Thibault and Yadlowsky, Steve and Tenney, Ian and Wei, Jason and Saphra, Naomi and D'Amour, Alexander and Linzen, Tal and Bastings, Jasmijn and Turc, Iulia Raluca and Eisenstein, Jacob and others},
    booktitle = {International Conference on Learning Representations},
    year      = {2021}
}

@inproceedings{blevins2022analyzing,
    title     = {Analyzing the Mono-and Cross-Lingual Pretraining Dynamics of Multilingual Language Models},
    author    = {Blevins, Terra and Gonen, Hila and Zettlemoyer, Luke},
    booktitle = {Proceedings of the 2022 Conference on Empirical Methods in Natural Language Processing},
    pages     = {3575--3590},
    year      = {2022}
}

@inproceedings{biderman2023pythia,
    title        = {Pythia: A suite for analyzing large language models across training and scaling},
    author       = {Biderman, Stella and Schoelkopf, Hailey and Anthony, Quentin Gregory and Bradley, Herbie and O’Brien, Kyle and Hallahan, Eric and Khan, Mohammad Aflah and Purohit, Shivanshu and Prashanth, USVSN Sai and Raff, Edward and others},
    booktitle    = {International Conference on Machine Learning},
    pages        = {2397--2430},
    year         = {2023},
    organization = {PMLR}
}

@inproceedings{xia2023training,
    title     = {Training Trajectories of Language Models Across Scales},
    author    = {Xia, Mengzhou and Artetxe, Mikel and Zhou, Chunting and Lin, Xi Victoria and Pasunuru, Ramakanth and Chen, Danqi and Zettlemoyer, Luke and Stoyanov, Ves},
    booktitle = {Proceedings of the 61st Annual Meeting of the Association for Computational Linguistics (Volume 1: Long Papers)},
    pages     = {13711--13738},
    year      = {2023}
}

@article{chang2023characterizing,
    title   = {Characterizing Learning Curves During Language Model Pre-Training: Learning, Forgetting, and Stability},
    author  = {Chang, Tyler A and Tu, Zhuowen and Bergen, Benjamin K},
    journal = {arXiv preprint arXiv:2308.15419},
    year    = {2023}
}

@article{chang2022word,
    title   = {Word acquisition in neural language models},
    author  = {Chang, Tyler A and Bergen, Benjamin K},
    journal = {Transactions of the Association for Computational Linguistics},
    volume  = {10},
    pages   = {1--16},
    year    = {2022}
}

@inproceedings{evanson2023language,
    title     = {Language acquisition: do children and language models follow similar learning stages?},
    author    = {Evanson, Linnea and Lakretz, Yair and King, Jean-R{\'e}mi},
    booktitle = {Findings of the Association for Computational Linguistics: ACL 2023},
    year      = {2023},
    pages     = {12205--12218}
}

@inproceedings{ma2025babysit,
    title     = {Babysit A Language Model From Scratch: Interactive Language Learning by Trials and Demonstrations},
    author    = {Ma, Ziqiao and Wang, Zekun and Chai, Joyce},
    booktitle = {Proceedings of the 2025 Conference of the Nations of the Americas Chapter of the Association for Computational Linguistics: Human Language Technologies (Volume 1: Long Papers)},
    pages     = {991--1010},
    year      = {2025}
}

@article{radford2019language,
    title   = {Language models are unsupervised multitask learners},
    author  = {Radford, Alec and Wu, Jeffrey and Child, Rewon and Luan, David and Amodei, Dario and Sutskever, Ilya and others},
    journal = {OpenAI blog},
    volume  = {1},
    number  = {8},
    pages   = {9},
    year    = {2019}
}

@inproceedings{kazemzadeh2014referitgame,
    address   = {Doha, Qatar},
    author    = {Kazemzadeh, Sahar  and
                 Ordonez, Vicente  and
                 Matten, Mark  and
                 Berg, Tamara},
    booktitle = {Proceedings of the 2014 Conference on Empirical Methods in Natural Language Processing ({EMNLP})},
    month     = {October},
    pages     = {787--798},
    publisher = {Association for Computational Linguistics},
    title     = {{R}efer{I}t{G}ame: Referring to Objects in Photographs of Natural Scenes},
    year      = {2014}
}

@article{mandler1992build,
    author    = {Mandler, Jean M},
    journal   = {Psychological review},
    number    = {4},
    pages     = {587},
    publisher = {American Psychological Association},
    title     = {How to build a baby: II. Conceptual primitives.},
    volume    = {99},
    year      = {1992}
}

@inproceedings{yu2016modeling,
    author       = {Yu, Licheng and Poirson, Patrick and Yang, Shan and Berg, Alexander C and Berg, Tamara L},
    booktitle    = {European Conference on Computer Vision},
    organization = {Springer},
    pages        = {69--85},
    title        = {Modeling context in referring expressions},
    year         = {2016}
}

@article{warstadt2022artificial,
    title     = {What artificial neural networks can tell us about human language acquisition},
    author    = {Warstadt, Alex and Bowman, Samuel R},
    journal   = {Algebraic Structures in Natural Language},
    pages     = {17--60},
    year      = {2022},
    publisher = {CRC Press}
}

@inproceedings{portelance2020predicting,
    title     = {Predicting Age of Acquisition in Early Word Learning Using Recurrent Neural Networks.},
    author    = {Portelance, Eva and Degen, Judith and Frank, Michael C},
    booktitle = {CogSci},
    year      = {2020}
}

@article{baroni2022proper,
    title     = {On the proper role of linguistically-oriented deep net analysis in linguistic theorizing},
    author    = {Baroni, Marco},
    journal   = {Algebraic structures in natural language},
    pages     = {1--16},
    year      = {2022},
    publisher = {Taylor and Francis}
}

@phdthesis{portelance2022neural,
    title  = {Neural Network Approaches to the Study of Word Learning},
    author = {Portelance, Eva},
    year   = {2022},
    school = {Stanford University}
}

@article{portelance2023roles,
    title   = {The roles of neural networks in language acquisition},
    author  = {Portelance, Eva and Jasbi, Masoud},
    journal = {PsyArXiv},
    year    = {2023}
}

@inproceedings{yang2025cambrians,
    title     = {Cambrian-s: Towards spatial supersensing in video},
    author    = {Yang, Shusheng and Yang, Jihan and Huang, Pinzhi and Brown II, Ellis L and Yang, Zihao and Yu, Yue and Tong, Shengbang and Zheng, Zihan and Xu, Yifan and Wang, Muhan and others},
    booktitle = {The Fourteenth International Conference on Learning Representations},
    year      = {2025}
}

@inproceedings{zhang2025do,
    title     = {Do Vision-Language Models Represent Space and How? Evaluating Spatial Frame of Reference under Ambiguities},
    author    = {Zheyuan Zhang and Fengyuan Hu and Jayjun Lee and Freda Shi and Parisa Kordjamshidi and Joyce Chai and Ziqiao Ma},
    booktitle = {The Thirteenth International Conference on Learning Representations},
    year      = {2025}
}

@inproceedings{gao2025vision,
    title     = {Do vision-language models have internal world models? towards an atomic evaluation},
    author    = {Qiyue Gao and Xinyu Pi and Kevin Liu and Junrong Chen and Ruolan Yang and Xinqi Huang and Xinyu Fang and Lu Sun and Gautham Kishore and Bo Ai and Stone Tao and Mengyang Liu and Jiaxi Yang and Chao-Jung Lai and Chuanyang Jin and Jiannan Xiang and Benhao Huang and Zeming Chen and David Danks and Hao Su and Tianmin Shu and Ziqiao Ma and Lianhui Qin and Zhiting Hu},
    booktitle = {Findings of the Association for Computational Linguistics: ACL 2025},
    year      = {2025}
}

@article{ha2018world,
    title   = {{World models}},
    author  = {Ha, David and Schmidhuber, J{\"u}rgen},
    journal = {arXiv preprint arXiv:1803.10122},
    year    = {2018}
}

@misc{fu2024blink,
    title         = {BLINK: Multimodal Large Language Models Can See but Not Perceive},
    author        = {Xingyu Fu and Yushi Hu and Bangzheng Li and Yu Feng and Haoyu Wang and Xudong Lin and Dan Roth and Noah A. Smith and Wei-Chiu Ma and Ranjay Krishna},
    year          = {2024},
    eprint        = {2404.12390},
    archiveprefix = {arXiv},
    primaryclass  = {cs.CV}
}

@misc{yi2020clevrer,
    title         = {CLEVRER: CoLlision Events for Video REpresentation and Reasoning},
    author        = {Kexin Yi and Chuang Gan and Yunzhu Li and Pushmeet Kohli and Jiajun Wu and Antonio Torralba and Joshua B. Tenenbaum},
    year          = {2020},
    eprint        = {1910.01442},
    archiveprefix = {arXiv},
    primaryclass  = {cs.CV}
}

@techreport{Gemini2023,
    title       = {Gemini: A Family of Highly Capable Multimodal Models},
    author      = {{Google}},
    year        = {2023},
    institution = {Google}
}

@book{johnson1983mental,
    title     = {Mental Models: Towards a Cognitive Science of Language, Inference, and Consciousness},
    author    = {Johnson-Laird, P.N.},
    isbn      = {9780674568822},
    lccn      = {83004333},
    series    = {Cognitive science series},
    url       = {https://books.google.com/books?id=FS3zSKAfLGMC},
    year      = {1983},
    publisher = {Harvard University Press}
}

@book{russell2016artificial,
    title     = {Artificial intelligence: a modern approach},
    author    = {Russell, Stuart J and Norvig, Peter},
    year      = {2016},
    publisher = {Pearson}
}

@misc{bear2022physion,
    title         = {Physion: Evaluating Physical Prediction from Vision in Humans and Machines},
    author        = {Daniel M. Bear and Elias Wang and Damian Mrowca and Felix J. Binder and Hsiao-Yu Fish Tung and R. T. Pramod and Cameron Holdaway and Sirui Tao and Kevin Smith and Fan-Yun Sun and Li Fei-Fei and Nancy Kanwisher and Joshua B. Tenenbaum and Daniel L. K. Yamins and Judith E. Fan},
    year          = {2022},
    eprint        = {2106.08261},
    archiveprefix = {arXiv},
    primaryclass  = {cs.AI}
}

@misc{bakhtin2019phyrenewbenchmarkphysical,
    title         = {PHYRE: A New Benchmark for Physical Reasoning},
    author        = {Anton Bakhtin and Laurens van der Maaten and Justin Johnson and Laura Gustafson and Ross Girshick},
    year          = {2019},
    eprint        = {1908.05656},
    archiveprefix = {arXiv},
    primaryclass  = {cs.LG},
    url           = {https://arxiv.org/abs/1908.05656}
}

@misc{shangguan2024tomatoassessingvisualtemporal,
    title         = {TOMATO: Assessing Visual Temporal Reasoning Capabilities in Multimodal Foundation Models},
    author        = {Ziyao Shangguan and Chuhan Li and Yuxuan Ding and Yanan Zheng and Yilun Zhao and Tesca Fitzgerald and Arman Cohan},
    year          = {2024},
    eprint        = {2410.23266},
    archiveprefix = {arXiv},
    primaryclass  = {cs.CV},
    url           = {https://arxiv.org/abs/2410.23266}
}

@article{lecun2022path,
    title   = {A path towards autonomous machine intelligence version 0.9. 2, 2022-06-27},
    author  = {LeCun, Yann},
    journal = {Open Review},
    volume  = {62},
    number  = {1},
    pages   = {1--62},
    year    = {2022}
}

@inproceedings{goyal2020rel3dminimallycontrastivebenchmark,
    author    = {Goyal, Ankit and Yang, Kaiyu and Yang, Dawei and Deng, Jia},
    booktitle = {Advances in Neural Information Processing Systems},
    editor    = {H. Larochelle and M. Ranzato and R. Hadsell and M.F. Balcan and H. Lin},
    pages     = {10514--10525},
    publisher = {Curran Associates, Inc.},
    title     = {Rel3D: A Minimally Contrastive Benchmark for Grounding Spatial Relations in 3D},
    url       = {https://proceedings.neurips.cc/paper_files/paper/2020/file/76dc611d6ebaafc66cc0879c71b5db5c-Paper.pdf},
    volume    = {33},
    year      = {2020}
}

@article{kappes2016mental,
    title     = {Mental simulation as substitute for experience},
    author    = {Kappes, Heather Barry and Morewedge, Carey K},
    journal   = {Social and Personality Psychology Compass},
    volume    = {10},
    number    = {7},
    pages     = {405--420},
    year      = {2016},
    publisher = {Wiley Online Library}
}

@article{wooldridge1995intelligent,
    title     = {Intelligent agents: Theory and practice},
    author    = {Wooldridge, Michael and Jennings, Nicholas R},
    journal   = {The knowledge engineering review},
    volume    = {10},
    number    = {2},
    pages     = {115--152},
    year      = {1995},
    publisher = {Cambridge University Press}
}

@inproceedings{nersessian1993theoretician,
    title        = {In the theoretician's laboratory: Thought experimenting as mental modeling},
    author       = {Nersessian, Nancy J},
    booktitle    = {PSA: Proceedings of the Biennial Meeting of the Philosophy of Science Association},
    volume       = {1992},
    number       = {2},
    pages        = {291--301},
    year         = {1993},
    organization = {Cambridge University Press}
}

@article{arunachalam2011grammatical,
    title     = {Grammatical form and semantic context in verb learning},
    author    = {Arunachalam, Sudha and Waxman, Sandra R},
    journal   = {Language Learning and Development},
    volume    = {7},
    number    = {3},
    pages     = {169--184},
    year      = {2011},
    publisher = {Taylor \& Francis}
}

@article{ferguson2014infants,
    title     = {Infants use known verbs to learn novel nouns: Evidence from 15-and 19-month-olds},
    author    = {Ferguson, Brock and Graf, Eileen and Waxman, Sandra R},
    journal   = {Cognition},
    volume    = {131},
    number    = {1},
    pages     = {139--146},
    year      = {2014},
    publisher = {Elsevier}
}

@incollection{cohen1999perceptual,
    author    = {L. Cohen and G. Amsel and M. A. Redford and M. Casasola},
    title     = {The Development of Infant Causal Perception},
    booktitle = {Perceptual development: Visual, auditory and speech perception in infancy},
    publisher = {Psychology Press},
    year      = {1999},
    editor    = {A. Slater}
}

@article{solstad2017causality,
    title   = {Causality and causal reasoning in natural language},
    author  = {Solstad, Torgrim and Bott, Oliver},
    journal = {The Oxford Handbook of Causal Reasoning},
    pages   = {619--644},
    year    = {2017}
}

@article{koppula2013learning,
    title     = {Learning Human Activities and Object Affordances from RGB-D videos},
    author    = {Koppula, Hema Swetha and Gupta, Rudhir and Saxena, Ashutosh},
    journal   = {The International Journal of Robotics Research},
    year      = {2013},
    number    = {8},
    pages     = {951--970},
    volume    = {32},
    publisher = {SAGE Publications}
}

@inproceedings{fathi2013modeling,
    title     = {Modeling Actions through State Changes},
    author    = {Fathi, Alireza and Rehg, James M},
    booktitle = {Proceedings of the IEEE Conference on Computer Vision and Pattern Recognition},
    pages     = {2579--2586},
    year      = {2013}
}

@inproceedings{yang2013detection,
    title        = {Detection of Manipulation Action Consequences (MAC)},
    author       = {Yang, Yezhou and Ferm{\"u}ller, Cornelia and Aloimonos, Yiannis},
    booktitle    = {Computer Vision and Pattern Recognition (CVPR), 2013 IEEE Conference on},
    pages        = {2563--2570},
    year         = {2013},
    organization = {IEEE}
}

@inproceedings{wang2016actions,
    title     = {Actions\~{} transformations},
    author    = {Wang, Xiaolong and Farhadi, Ali and Gupta, Abhinav},
    booktitle = {Proceedings of the IEEE conference on Computer Vision and Pattern Recognition},
    pages     = {2658--2667},
    year      = {2016}
}

@article{zhuo2019explainable,
    title   = {Explainable Video Action Reasoning via Prior Knowledge and State Transitions},
    author  = {Zhuo, Tao and Cheng, Zhiyong and Zhang, Peng and Wong, Yongkang and Kankanhalli, Mohan},
    journal = {ACM Multimedia},
    year    = {2019}
}

@article{fire2015learning,
    title     = {Learning Perceptual Causality from Video},
    author    = {Fire, Amy and Zhu, Song-Chun},
    journal   = {ACM Transactions on Intelligent Systems and Technology (TIST)},
    year      = {2015},
    number    = {2},
    pages     = {23},
    volume    = {7},
    publisher = {ACM}
}

@inproceedings{fire2017inferring,
    title     = {Inferring Hidden Statuses and Actions in Video by Causal Reasoning},
    author    = {Fire, Amy. and Zhu, Song-Chun.},
    year      = {2017},
    booktitle = {CVPR Workshop: Vision Meets Cognition}
}

@inproceedings{alayrac2017joint,
    title     = {Joint Discovery of Object States and Manipulation Actions},
    author    = {Alayrac, Jean-Baptiste and Laptev, Ivan and Sivic, Josef and Lacoste-Julien, Simon},
    booktitle = {Proceedings of the IEEE International Conference on Computer Vision},
    pages     = {2127--2136},
    year      = {2017}
}

@inproceedings{liu2017joint,
    author    = {Liu, Yang and Wei, Ping and Zhu, Song-Chun},
    booktitle = {2017 IEEE International Conference on Computer Vision (ICCV)},
    title     = {Jointly Recognizing Object Fluents and Tasks in Egocentric Videos},
    year      = {2017},
    pages     = {2943-2951}
}

@inproceedings{hong2021transform,
    author    = {Hong, Xin and Lan, Yanyan and Pang, Liang and Guo, Jiafeng and Cheng, Xueqi},
    title     = {Transformation Driven Visual Reasoning},
    booktitle = {Proceedings of the IEEE/CVF Conference on Computer Vision and Pattern Recognition (CVPR)},
    month     = {June},
    year      = {2021},
    pages     = {6903-6912}
}

@inproceedings{zhou2016learning,
    title        = {Learning temporal transformations from time-lapse videos},
    author       = {Zhou, Yipin and Berg, Tamara L},
    booktitle    = {European Conference on Computer Vision},
    pages        = {262--277},
    year         = {2016},
    organization = {Springer}
}

@inproceedings{xiong2018learning,
    author    = {Xiong, Wei and Luo, Wenhan and Ma, Lin and Liu, Wei and Luo, Jiebo},
    title     = {Learning to Generate Time-Lapse Videos Using Multi-Stage Dynamic
                 Generative Adversarial Networks},
    booktitle = {The IEEE Conference on Computer Vision and Pattern Recognition
                 (CVPR)},
    month     = {June},
    year      = {2018}
}

@inproceedings{gao2018action,
    title     = {What Action Causes This? Towards Naive Physical Action-Effect Prediction},
    author    = {Gao, Qiaozi  and
                 Yang, Shaohua  and
                 Chai, Joyce  and
                 Vanderwende, Lucy},
    booktitle = {Proceedings of the 56th Annual Meeting of the Association for Computational Linguistics (Volume 1: Long Papers)},
    month     = jul,
    year      = {2018},
    address   = {Melbourne, Australia},
    publisher = {Association for Computational Linguistics},
    pages     = {934--945}
}

@inproceedings{wu2015galileo,
    title     = {Galileo: Perceiving physical object properties by integrating a physics engine with deep learning},
    author    = {Wu, Jiajun and Yildirim, Ilker and Lim, Joseph J and Freeman, William T and Tenenbaum, Joshua B},
    booktitle = {Advances in Neural Information Processing Systems},
    pages     = {127--135},
    year      = {2015}
}

@inproceedings{wu2016physics,
    title     = {Physics 101: Learning Physical Object Properties from Unlabeled Videos},
    author    = {Wu, Jiajun and Lim, Joseph J and Zhang, Hongyi and Tenenbaum, Joshua B and Freeman, William T},
    booktitle = {BMVC},
    volume    = {2},
    number    = {6},
    pages     = {7},
    year      = {2016}
}

@inproceedings{wu2017learning,
    title     = {Learning to see physics via visual de-animation},
    author    = {Wu, Jiajun and Lu, Erika and Kohli, Pushmeet and Freeman, Bill and Tenenbaum, Josh},
    booktitle = {Advances in Neural Information Processing Systems},
    pages     = {152--163},
    year      = {2017}
}

@article{gentner1982nouns,
    title     = {Why nouns are learned before verbs: Linguistic relativity versus natural partitioning},
    author    = {Gentner, Dedre},
    journal   = {Center for the Study of Reading Technical Report; no. 257},
    year      = {1982},
    publisher = {Champaign, Ill.: University of Illinois at Urbana-Champaign, Center for the~…}
}

@article{spelke1990principles,
    title     = {Principles of object perception},
    author    = {Spelke, Elizabeth S},
    journal   = {Cognitive science},
    volume    = {14},
    number    = {1},
    pages     = {29--56},
    year      = {1990},
    publisher = {Wiley Online Library}
}

@article{gentner2006verbs,
    title     = {Why verbs are hard to learn},
    author    = {Gentner, Dedre},
    journal   = {Action meets word: How children learn verbs},
    pages     = {544--564},
    year      = {2006},
    publisher = {Citeseer}
}

@article{cohen1998development,
    title   = {The development of infant causal perception},
    author  = {Cohen, Leslie B and Amsel, Geoffrey and Redford, Melissa A and Casasola, Marianella},
    journal = {Perceptual development: Visual, auditory, and speech perception in infancy},
    pages   = {167--209},
    year    = {1998}
}

@article{leslie1987six,
    title     = {Do six-month-old infants perceive causality?},
    author    = {Leslie, Alan M and Keeble, Stephanie},
    journal   = {Cognition},
    volume    = {25},
    number    = {3},
    pages     = {265--288},
    year      = {1987},
    publisher = {Elsevier}
}

@article{oakes1990infant,
    title     = {Infant perception of a causal event},
    author    = {Oakes, Lisa M and Cohen, Leslie B},
    journal   = {Cognitive Development},
    volume    = {5},
    number    = {2},
    pages     = {193--207},
    year      = {1990},
    publisher = {Elsevier}
}

@article{golinkoff1987eyes,
    title     = {The eyes have it: Lexical and syntactic comprehension in a new paradigm},
    author    = {Golinkoff, Roberta Michnick and Hirsh-Pasek, Kathryn and Cauley, Kathleen M and Gordon, Laura},
    journal   = {Journal of child language},
    volume    = {14},
    number    = {1},
    pages     = {23--45},
    year      = {1987},
    publisher = {Cambridge University Press}
}

@inproceedings{buettner2024investigating,
    title     = {Investigating the role of attribute context in vision-language models for object recognition and detection},
    author    = {Buettner, Kyle and Kovashka, Adriana},
    booktitle = {Proceedings of the IEEE/CVF Winter Conference on Applications of Computer Vision},
    pages     = {5474--5484},
    year      = {2024}
}

@article{levin2008constraint,
    title   = {A constraint on verb meanings: Manner/result complementarity},
    author  = {Levin, Beth},
    journal = {Cognitive Science Department Colloqium Series, Brown University, Providence, RI, March},
    volume  = {17},
    pages   = {2008},
    year    = {2008}
}

@inproceedings{sener2022assembly101,
    title     = {Assembly101: A large-scale multi-view video dataset for understanding procedural activities},
    author    = {Sener, Fadime and Chatterjee, Dibyadip and Shelepov, Daniel and He, Kun and Singhania, Dipika and Wang, Robert and Yao, Angela},
    booktitle = {Proceedings of the IEEE/CVF Conference on Computer Vision and Pattern Recognition},
    pages     = {21096--21106},
    year      = {2022}
}

@misc{grauman2022ego4dworld3000hours,
    title         = {Ego4D: Around the World in 3,000 Hours of Egocentric Video},
    author        = {Kristen Grauman and Andrew Westbury and Eugene Byrne and Zachary Chavis and Antonino Furnari and Rohit Girdhar and Jackson Hamburger and Hao Jiang and Miao Liu and Xingyu Liu and Miguel Martin and Tushar Nagarajan and Ilija Radosavovic and Santhosh Kumar Ramakrishnan and Fiona Ryan and Jayant Sharma and Michael Wray and Mengmeng Xu and Eric Zhongcong Xu and Chen Zhao and Siddhant Bansal and Dhruv Batra and Vincent Cartillier and Sean Crane and Tien Do and Morrie Doulaty and Akshay Erapalli and Christoph Feichtenhofer and Adriano Fragomeni and Qichen Fu and Abrham Gebreselasie and Cristina Gonzalez and James Hillis and Xuhua Huang and Yifei Huang and Wenqi Jia and Weslie Khoo and Jachym Kolar and Satwik Kottur and Anurag Kumar and Federico Landini and Chao Li and Yanghao Li and Zhenqiang Li and Karttikeya Mangalam and Raghava Modhugu and Jonathan Munro and Tullie Murrell and Takumi Nishiyasu and Will Price and Paola Ruiz Puentes and Merey Ramazanova and Leda Sari and Kiran Somasundaram and Audrey Southerland and Yusuke Sugano and Ruijie Tao and Minh Vo and Yuchen Wang and Xindi Wu and Takuma Yagi and Ziwei Zhao and Yunyi Zhu and Pablo Arbelaez and David Crandall and Dima Damen and Giovanni Maria Farinella and Christian Fuegen and Bernard Ghanem and Vamsi Krishna Ithapu and C. V. Jawahar and Hanbyul Joo and Kris Kitani and Haizhou Li and Richard Newcombe and Aude Oliva and Hyun Soo Park and James M. Rehg and Yoichi Sato and Jianbo Shi and Mike Zheng Shou and Antonio Torralba and Lorenzo Torresani and Mingfei Yan and Jitendra Malik},
    year          = {2022},
    eprint        = {2110.07058},
    archiveprefix = {arXiv},
    primaryclass  = {cs.CV},
    url           = {https://arxiv.org/abs/2110.07058}
}

@article{Damen2020RESCALING,
    title   = {Rescaling Egocentric Vision},
    author  = {Damen, Dima and Doughty, Hazel and Farinella, Giovanni Maria  and and Furnari, Antonino and Ma, Jian and Kazakos, Evangelos and Moltisanti, Davide and Munro, Jonathan and Perrett, Toby and Price, Will and Wray, Michael},
    journal = {CoRR},
    volume  = {abs/2006.13256},
    year    = {2020},
    ee      = {http://arxiv.org/abs/2006.13256}
}

@inproceedings{Damen2018EPICKITCHENS,
    title     = {Scaling Egocentric Vision: The EPIC-KITCHENS Dataset},
    author    = {Damen, Dima and Doughty, Hazel and Farinella, Giovanni Maria  and Fidler, Sanja and
                 Furnari, Antonino and Kazakos, Evangelos and Moltisanti, Davide and Munro, Jonathan
                 and Perrett, Toby and Price, Will and Wray, Michael},
    booktitle = {European Conference on Computer Vision (ECCV)},
    year      = {2018}
}

@article{liu2022fineaction,
    author   = {Liu, Yi and Wang, Limin and Wang, Yali and Ma, Xiao and Qiao, Yu},
    journal  = {IEEE Transactions on Image Processing},
    title    = {FineAction: A Fine-Grained Video Dataset for Temporal Action Localization},
    year     = {2022},
    volume   = {31},
    number   = {},
    pages    = {6937-6950},
    doi      = {10.1109/TIP.2022.3217368}
}

@inproceedings{HoloAssist2023,
    author    = {Wang, Xin and Kwon, Taein and Rad, Mahdi and Pan, Bowen and Chakraborty, Ishani and Andrist, Sean and Bohus, Dan and Feniello, Ashley and Tekin, Bugra and Frujeri, Felipe Vieira and Joshi, Neel and Pollefeys, Marc},
    title     = {HoloAssist: an Egocentric Human Interaction Dataset for Interactive AI Assistants in the Real World},
    booktitle = {Proceedings of the IEEE/CVF International Conference on Computer Vision (ICCV)},
    month     = {October},
    year      = {2023},
    pages     = {20270-20281}
}

@misc{flux-2-2025,
    author       = {Black Forest Labs},
    title        = {{FLUX.2: Frontier Visual Intelligence}},
    year         = {2025},
    howpublished = {\url{https://bfl.ai/blog/flux-2}}
}

@article{xu2025next,
    title   = {Next-Embedding Prediction Makes Strong Vision Learners},
    author  = {Xu, Sihan and Ma, Ziqiao and Chai, Wenhao and Chen, Xuweiyi and Jin, Weiyang and Chai, Joyce and Xie, Saining and Yu, Stella X},
    journal = {arXiv preprint arXiv:2512.16922},
    year    = {2025}
}

@article{baker2009action,
  title={Action understanding as inverse planning},
  author={Baker, Chris L and Saxe, Rebecca and Tenenbaum, Joshua B},
  journal={Cognition},
  volume={113},
  number={3},
  pages={329--349},
  year={2009},
  publisher={Elsevier}
}

@inproceedings{gao2018causality,
 author = {Qiaozi Gao and
Malcolm Doering and
Shaohua Yang and
Joyce Chai},
 bibsource = {dblp computer science bibliography, https://dblp.org},
 booktitle = {Proceedings of the 54th Annual Meeting of the Association for Computational
Linguistics, ACL 2016, August 7-12, 2016, Berlin, Germany, Volume
1: Long Papers},
 doi = {10.18653/v1/p16-1171},
 publisher = {The Association for Computer Linguistics},
 title = {Physical Causality of Action Verbs in Grounded Language Understanding},
 url = {https://doi.org/10.18653/v1/p16-1171},
 year = {2016}
}

@article{leslie-infant,
title = {Do six-month-old infants perceive causality?},
journal = {Cognition},
volume = {25},
number = {3},
pages = {265-288},
year = {1987},
issn = {0010-0277},
doi = {https://doi.org/10.1016/S0010-0277(87)80006-9},
url = {https://www.sciencedirect.com/science/article/pii/S0010027787800069},
author = {Alan M. Leslie and Stephanie Keeble}
}

@article{leslie-infant-perception,
author = {Leslie B. Cohen and Leslie J. Rundell and Barbara A. Spellman and Cara H. Cashon},
title ={Infants' Perception of Causal Chains},
journal = {Psychological Science},
volume = {10},
number = {5},
pages = {412-418},
year = {1999},
doi = {10.1111/1467-9280.00178},
URL = {https://doi.org/10.1111/1467-9280.00178},
eprint = {https://doi.org/10.1111/1467-9280.00178}
}
